\documentclass[11pt]{article}
 
\usepackage{setspace}
\usepackage[vmargin=1in,hmargin=1in]{geometry}

\usepackage{authblk}

\usepackage{amsmath,amssymb}
\usepackage{graphicx}
\usepackage{caption}
\DeclareCaptionLabelSeparator{line}{ $\mid$ }
\usepackage[colorlinks=true,bookmarks=false,citecolor=blue,urlcolor=blue]{hyperref} 
\usepackage{siunitx}
\usepackage{amsmath}
\usepackage{float}
\usepackage[justification=justified,singlelinecheck=false]{caption}

\usepackage[superscript,nomove]{cite}

\usepackage[acronym,shortcuts]{glossaries}
\newacronym{AI}{AI}{artificial intelligence}
\newacronym{ONN}{ONN}{optical neural network}
\newacronym{PIC}{PIC}{photonic integrated circuit}
\newacronym{MVM}{MVM}{matrix-vector multiplication}
\newacronym{ASIC}{ASIC}{application-specific integrated circuit}
\newacronym{TPU}{TPU}{tensor processing unit}
\newacronym{DAC}{DAC}{digital-to-analogue conversion}
\newacronym{DOCP}{DOCP}{digital optical convolution processor}
\newacronym{DSP}{DSP}{digital signal processing}
\newacronym{MRM}{MRM}{microring modulator}
\newacronym{RF}{RF}{radio frequency}
\newacronym{WDM}{WDM}{wavelength-division multiplexing}
\newacronym{OOK}{OOK}{on-off keying}
\newacronym{RMSE}{RMSE}{root-mean-square error}
\newacronym{CNN}{CNN}{convolutional neural network}
\newacronym{SNR}{SNR}{signal-to-noise ratio}
\newacronym{PER}{PER}{pixel error rate}
\newacronym{PD}{PD}{photodiode}
\newacronym{APD}{APD}{avalanche photodiode}
\newacronym{FPGA}{FPGA}{field programmable gate arrays}
\newacronym{PAM}{PAM}{pulse amplitude modulation}
\newacronym{ADC}{ADC}{analogue-to-digital conversion}
\newacronym{OSNR}{OSNR}{optical signal-to-noise ratio}
\newacronym{OSA}{OSA}{optical spectrum analyzer}
\newacronym{MAC}{MAC}{Multiply–accumulate operation}
\newacronym{HDIP}{HDIP}{high-definition image processing}
\newacronym{HWDR}{HWDR}{handwritten digit recognition}
\newacronym{AlGaAs}{AlGaAs}{aluminum gallium arsenide}
\newacronym{AlGaAsOI}{AlGaAsOI}{aluminum gallium arsenide on insulator}
\newacronym{FWM}{FWM}{four-wave mixing}
\newacronym{TPA}{TPA}{two-photon absorption}
\newacronym{CW}{CW}{continuous-wave}
\newacronym{QAM}{QAM}{quadrature-amplitude modulation}
\newacronym{N-WDM}{N-WDM}{Nyquist wavelength-division multiplexing}
\newacronym{CE}{CE}{conversion efficiency}
\newacronym{TDFA}{TDFA}{thulium-doped fibre amplifier}
\newacronym{TDF}{TDF}{thulium-doped fibre}
\newacronym{ZDW}{ZDW}{zero-dispersion wavelength}
\newacronym{DFB}{DFB}{distributed feedback}
\newacronym{HCF}{HCF}{hollow-core fibre}
\newacronym{SSMF}{SSMF}{standard single-mode fibre}
\newacronym{EDFA}{EDFA}{erbium-doped fibre amplifier}
\newacronym{BER}{BER}{bit error ratio}
\newacronym{LDPC}{LDPC}{low-density parity-check}
\newacronym{HD-FEC}{HD-FEC}{hard-decision forward error correction}
\newacronym{FEC}{FEC}{forward error correction}
\newacronym{DCF}{DCF}{dispersion compensating fibre}
\newacronym{ECL}{ECL}{external cavity laser}
\newacronym{DVB-S2}{DVB-S2}{digital video broadcasting – satellite – second generation}
\newacronym{OBPF}{OBPF}{optical bandpass fiter}
\newacronym{OFC}{OFC}{optical frequency comb}
\newacronym{DF-HNLF}{DF-HNLF}{dispersion-flattened highly nonlinear fibre}
\newacronym{GVD}{GVD}{group velocity dispersion}
\newacronym{WSS}{WSS}{wavelength-selective switch}
\newacronym{PCB}{PCB}{printed circuit board}
\newacronym{AWG}{AWG}{arbitrary waveform generator}
\newacronym{DSO}{DSO}{digital storage oscilloscope}
\newacronym{MZI}{MZI}{Mach-Zehnder interferometer}
\newacronym{HOP}{HOP}{hybrid optical processor}
\newacronym{ENOB}{ENOB}{effective number of bits}
\newacronym{MNIST}{MNIST}{modified National Institute of Standards and Technology}

\title{\textbf{Digital-analog hybrid matrix multiplication processor for optical neural networks}}

\author[1,$\dagger$]{Xiansong~Meng}
\author[1,$\dagger$,$\ast$]{Deming~Kong}
\author[2]{Kwangwoong~Kim}
\author[3]{Qiuchi~Li}
\author[4]{Po~Dong}
\author[3,5]{Ingemar~J.~Cox}
\author[3]{Christina~Lioma}
\author[1,$\ast$]{Hao~Hu}

\affil[1]{DTU Electro, Technical University of Denmark, DK-2800, Kgs. Lyngby, Denmark}
\affil[2]{Nokia Bell Labs, New Providence, NJ 07974, USA}
\affil[3]{Department of Computer Science, University of Copenhagen, Denmark}
\affil[4]{II-VI Incorporated, 48800 Milmont Dr., Fremont, CA 94538, USA}
\affil[5]{Department of Computer Science, University College London, UK}

\affil[$\dagger$]{These authors contributed equally}

\date{}

\begin{document}

\maketitle

\section*{Corresponding authors}
Correspondence to: \\
Deming Kong (dmkon@dtu.dk), ORCID: 0000-0001-6552-4081\\
Hao Hu (huhao@dtu.dk), ORCID: 0000-0002-8859-0986

\section*{Abstract}
The computational demands of modern AI have spurred interest in \glspl*{ONN} which offer the potential benefits of increased speed and lower power consumption. However, current \glspl*{ONN} face various challenges, most significantly a limited calculation precision (typically around 4 bits) and the requirement for high-resolution signal format converters (\glspl*{DAC} and \glspl*{ADC}). These challenges are inherent to their analog computing nature and pose significant obstacles in practical implementation. Here, we propose a digital-analog hybrid optical computing architecture for \glspl*{ONN}, which utilizes digital optical inputs in the form of binary words. By introducing the logic levels and decisions based on thresholding, the calculation precision can be significantly enhanced. The \glspl*{DAC} for input data can be removed and the resolution of the \glspl*{ADC} can be greatly reduced. This can increase the operating speed at a high calculation precision and facilitate the compatibility with microelectronics. To validate our approach, we have fabricated a proof-of-concept photonic chip and built up a \gls*{HOP} system for neural network applications. We have demonstrated an unprecedented 16-bit calculation precision for high-definition image processing, with a \gls*{PER} as low as $1.8\times10^{-3}$ at an \gls*{SNR} of 18.2 dB. We have also implemented a convolutional neural network for handwritten digit recognition that shows the same accuracy as the one achieved by a desktop computer. The concept of the digital-analog hybrid optical computing architecture offers a methodology that could potentially be applied to various \gls*{ONN} implementations and may intrigue new research into efficient and accurate domain-specific optical computing architectures for neural networks.  

\glsresetall

\section*{Main}
Modern artificial intelligence based on deep learning algorithms has demonstrated impressive capabilities \cite{wu2023brief}. However, these algorithms require enormous computing power and corresponding energy. The demand for computing power is now doubling every 3--4 months\cite{Amodei18}, a rate surpassing the well-known Moore's law. This has given rise to domain-specific hardware accelerators using \glspl*{ASIC}, for example, Google's \glspl*{TPU}\cite{jouppi17} and IBM's TrueNorth\cite{merolla14}. The aim is to develop an efficient hardware platform with advanced parallelism for matrix multiplications. However, microelectronics is encountering fundamental bottlenecks in speed, energy consumption, heating, and interconnect delay, which become increasingly hard to be resolved by scaling\cite{mooreslaw,rupp21,khan18}. 

\Glspl*{PIC} present a pathway free from these obstacles, and hence form a promising disruptive computing architecture beyond von Neumann architecture and Moore's Law to potentially accelerate neural network applications efficiently\cite{caulfield10NP,miller15Optica,peng18JSTQE}. Consequently, integrated \gls*{ONN} have been proposed to address the obstacles of microelectronics\cite{shen17NP,feldmann21Nature}, showing potential to surpass their digital microelectronic counterparts in calculating speed, energy consumption, as well as computing density\cite{nahmias19}. Despite the advantages of \glspl*{ONN}, the demonstrations are all based on analog computing architectures where the input and weight vectors (i.e., the multipliers and multiplicands for the matrix multiplication) are represented by light intensities. The analog computing nature of \glspl*{ONN} gives rise to a major scientific challenge: insufficient signal-to-noise ratio due to accumulated noise, and crosstalk in the computing system. This imposes several fundamental limitations. One primary limitation is the low precision of calculations. The intensity resolution of the optical signals is usually limited to a calculation precision around 4 bits\cite{tait16,tait17}. Efforts have recently been made to increase the control precision of the weight values up to 9 bits for some analog \gls*{ONN} schemes\cite{zhang22Optica,bai23NC}. However, this does not directly translate into a calculation precision of 9 bits for the matrix multiplication, but rather 4.2 bits based on a 99.7$\%$ confidence interval derived from the noise distribution of the obtained results\cite{bai23NC}. In practice, 16-bit calculation precision is required for reasonable training convergence and many demanding inference tasks such as autonomous driving, image processing, and 3D computer vision\cite{gupta15}. Second, analog optical computing is incompatible with microelectronics, requiring high-resolution \glspl*{DAC} and \glspl*{ADC} which are expensive and energy-consuming. 

Here, we propose a new type of digital-analog hybrid matrix multiplication processor for optical neural networks. The hybrid optical processor differs fundamentally from existing analog \gls*{ONN} processors and offers the following benefits. The introduction of logic levels can significantly increase the calculation precision for matrix multiplication. Powerful \gls*{DSP} algorithms can improve calculation performance and ensure high calculation repeatability. The high-resolution \glspl*{DAC} for the inputs can be removed; and the requirements for \glspl*{ADC} for the outputs can be greatly released by $M$ bits, considering \gls*{MVM} of $N$-bit multipliers and $M$-bit multiplicands. In return, this can increase the operating speed and improve the compatibility with microelectronics. We propose the concept of a \gls*{HOP} in convolution neural networks. Our simulations show good noise tolerance and improved performance of the \gls*{HOP} over analog optical computing scheme. At an \gls*{SNR} of 25 dB, our \gls*{HOP} can achieve an \gls*{RMSE} of $1.2\times10^{-3}$ for an 8-bit image processing task with a 3$\times$3 convolutional operator. We have also built a proof-of-concept silicon photonic chip and applied the \gls*{HOP} in a 16-bit depth \gls*{HDIP} system, as well as \gls*{HWDR}. The convolution results show that a record high calculation precision of 16 bits is successfully achieved with a \gls*{PER} of $\text{1.8}\times\text{10}^\text{-3}$ at a \gls*{SNR} of 18.2 dB and an operating speed of 7.5 GHz. The \gls*{PER} performance is also investigated under various optical noise levels for an 8-bit image processing task, showing a \gls*{PER} of $6.0\times10^{-5}$ at an \gls*{OSNR} of 35 dB. The \gls*{HWDR} shows the same accuracy as the one calculated with a desktop computer. Notably, the \gls*{HOP} is a method that could potentially be applied to other \gls*{ONN} schemes and spark new concepts considering domain-specific optical computing.

\section*{The principle of the \acrlong*{HOP}}

\begin{figure}[htbp]
    \centering
    \includegraphics[width=0.99\linewidth]{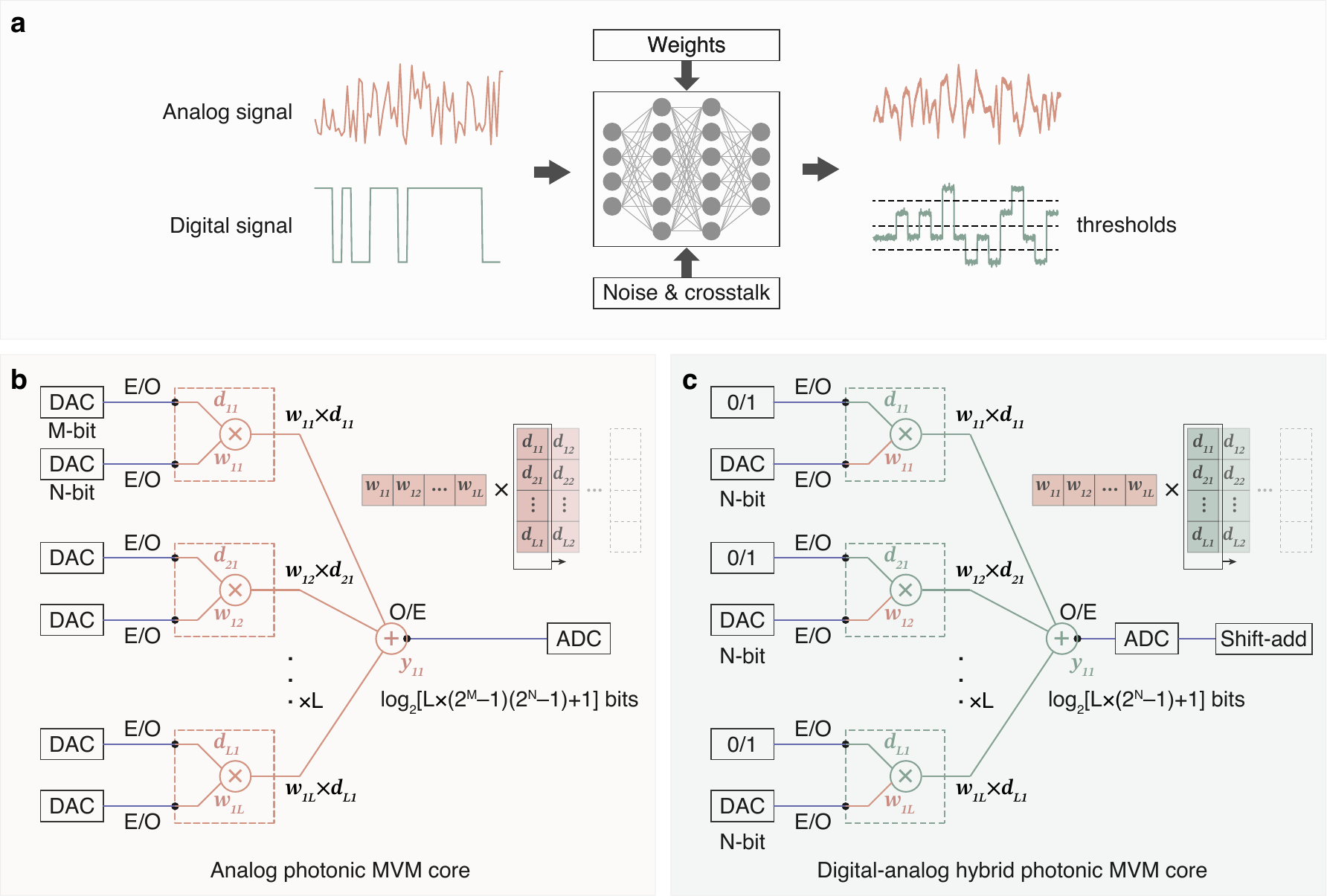}
    \caption{\textbf{Concept of the digital-analog hybrid photonic \acrshort*{MVM} core.} \textbf{a,} Digital signal is more robust to noise and crosstalk in a signal processing system. \Acrshort*{ONN} can be seen as a signal processing system where digital signals can potentially be applied for better calculation repeatability, precision, scalability, and compatibility with microelectronics compared to analog signals. \textbf{b,} The abstracted analog photonic \acrshort*{MVM} core utilizing analog signals for both input $d$ and weight $w$. \textbf{c,} The proposed digital-analog hybrid photonic \acrshort*{MVM} core utilizing digital signals for input $d$, with relaxed constraints for signal format converters.}
    \label{fig:concept}
\end{figure}

\noindent Analog signals are fundamentally vulnerable to noise and crosstalk. The optical computing system can be seen as an optical signal processing system and \gls*{ONN} is a representation of layers of the optical matrix multiplication system. Shown in Fig. \ref{fig:concept}a, analog signals traversing the signal processing system with noise and crosstalk suffer more severe degradation than digital signals under the same \gls*{SNR}. This is due to the absence of logic levels and decision-making processes in analog systems, which prevents the signal from being effectively recovered and equalized. The hypothesis here is that the performance of the optical computing system can benefit from the utilization of digital optical signals.

Analog optical matrix multiplication architectures rely on analog photonic multiplication cores, abstractly represented in Fig. \ref{fig:concept}b. The input $d$ and weight $w$ originate from M-bit and N-bit \glspl*{DAC} in the electrical domain. After electro-optical conversion, $d$ and $w$ are multiplied and converted back into the electrical domain. This generates an optical signal with $\left(2^M-1\right)\times\left(2^N-1\right)$ possible levels (considering signal levels starting from 0, thus $2^M\times2^N-2^M-2^N+1$). To accomplish a full-precision multiplication without losing information, an \gls*{ADC} with a resolution (\gls*{ENOB}) of $\text{log}_2\left[\left(2^M-1\right)\times\left(2^N-1\right)\right]$ bits is necessary. This can be very challenging, especially for high-speed operations, since speed-versus-resolution is a well-known trade-off for the signal converters \cite{adc_survey}. While it is possible to reduce the operational speed to attain the desired resolution, the high-speed operation advantage of photonic components vanishes. On the contrary, while high-speed operations are maintained, the trade-off is an inevitable reduction in calculation precision (Details are discussed in Supplementary Table 1, 2 and Supplementary Figure 2). The number of overall possible signal levels increases proportionally for calculating an inner product of an input vector and a weight vector, generating $L\times\left(2^M-1\right)\times\left(2^N-1\right)$ possible levels, which requires an \gls*{ADC} with a resolution of $\text{log}_{2}\left[L\times\left(2^M-1\right)\times\left(2^N-1\right)\right]$ bits. 

The \gls*{HOP} concept employs digital binary words for matrix multiplication into the optical computing system. For single \gls*{MVM} operation, $d$ is carried in digital optical signals using binary words, while $w$ is kept in analog format leaving the design of the computing system unchanged. If $w$ originates from an N-bit \gls*{DAC}, the multiplication result is a signal with a reduced potential number of $2^N-1$ levels, which requires an \gls*{ADC} resolution of $\text{log}_2\left(2^N-1\right)$ bits. The \gls*{HOP} concept can be extended to build a hybrid digital-analog photonic \gls*{MVM} core represented in Fig. \ref{fig:concept}c. An input vector $(d_{11}, d_{21}, ..., d_{L1})^T$ and a weight vector $(w_{11}, w_{12}, ..., w_{1L})$ can be input to $L$ multipliers. The results from the multipliers are kept in the optical domain and summed in a photo-detector. These are then processed by a post-processing electrical circuit to get the result of the vector inner product (see Supplementary Figure 1). Columns of the input matrix $\textbf{D}$ are loaded sequentially to accomplish the \gls*{MVM} operation. The vector multiplication yields $L\times\left(2^N-1\right)$ signal levels, requiring an \gls*{ADC} with a resolution of $\text{log}_{2}\left[L\times\left(2^N-1\right)\right]$ bits. 
Note that the resolution of the \gls*{ADC} is independent of the input vector $d$. In theory, the hybrid photonic \gls*{MVM} core can support input values of any precision without sacrificing the performance. 

By eliminating the need for a high-speed \gls*{DAC} in processing the input matrix $\textbf{D}$, the hybrid \gls*{MVM} core reduces costs, saves energy, and simplifies system complexity. The requirement for the resolution of \gls*{ADC} of the hybrid multiplier is $\sim$$M$-bit less than the analog multiplier, potentially increasing the operating speed (see Supplementary Table 1, 2 and Supplementary Figure 2). The relaxed constraints on ADC/DAC converters can improve the processor's compatibility with microelectronics. From an information encoding point of view, the analog \gls*{ONN} processors revolve around employing only the signal amplitude for encoding information, while the \gls*{HOP} utilizes both amplitude and time for information encoding. The increased encoding space in \gls*{HOP} results in a larger Euclidean distance between encoded signal samples when compared to analog \gls*{ONN} processors. As a result, the \gls*{HOP} has a better performance against noise and exhibit a higher calculation precision.

\section*{An implementation of the digital-analog hybrid photonic \gls*{MVM} core}

\begin{figure}[htbp]
    \centering
    \includegraphics[width=0.99\linewidth]{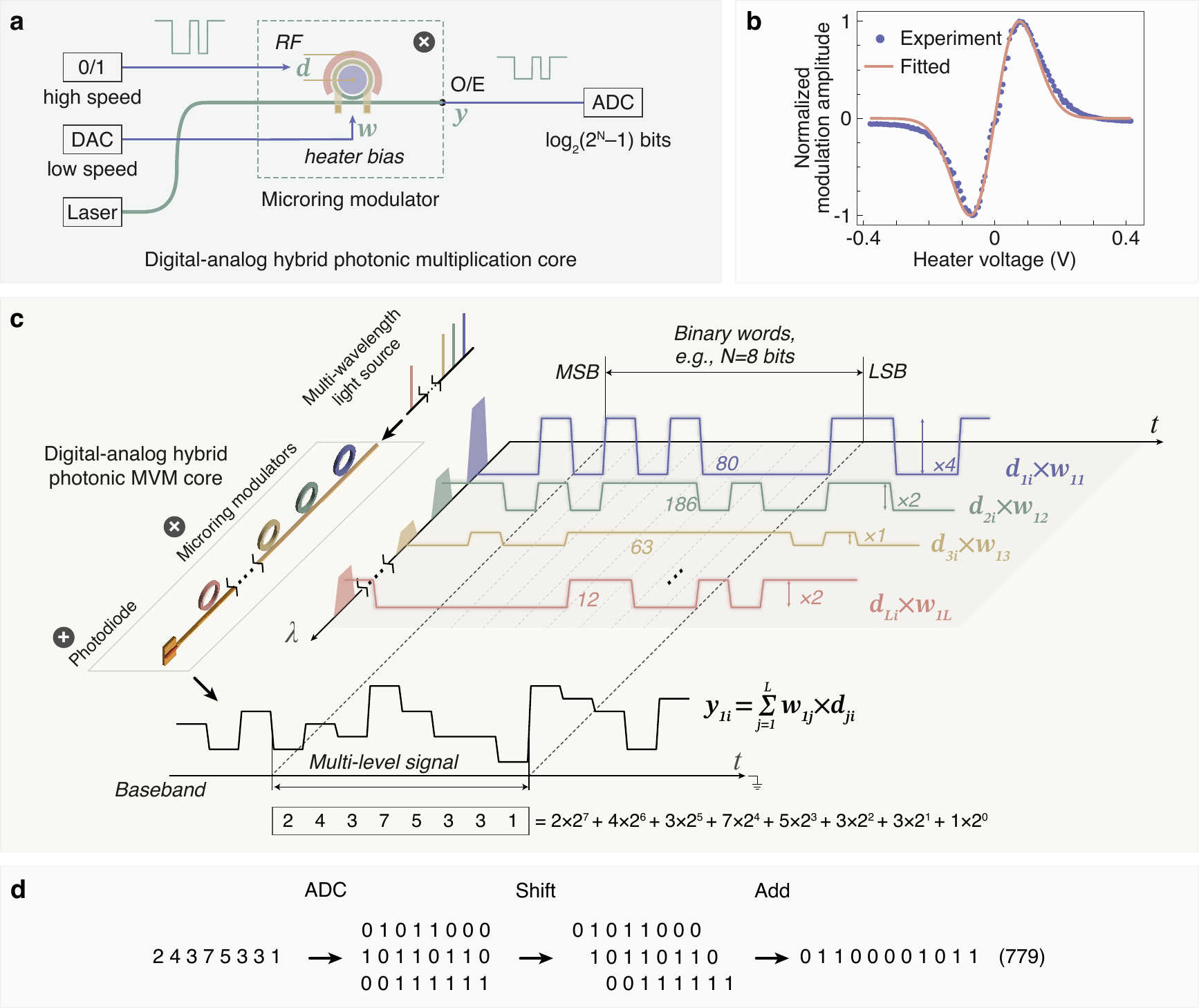}
    \caption{\textbf{A \acrshort*{MRM} based implementation of the hybrid digital-analog photonic \acrshort*{MVM} core.} \textbf{a,} The implementation of a single hybrid digital-analog photonic multiplication core using an \acrshort*{MRM}, where the input $d$ in the form of a digital optical signal is loaded through the high-speed port while the weight $w$ is loaded using microheater based modulation bias. \textbf{b,} The measured relationship between the normalized weight and the required heater voltage for the modulation bias, can be used as a lookup table to load the weight. \textbf{c,} The implementation and optical signal temporal evolution of the digital-analog hybrid photonic \acrshort*{MVM} core, including a multi-wavelength light source, an array of microring modulators, and a photodiode. \textbf{d,} Post-processing of the multilevel signal includes an \acrshort*{ADC} and a shift-add operation. Here the multilevel signal is converted to a binary signal and the final result can be recovered via shifts and adds.}
    \label{fig:principle}
\end{figure}

\noindent The \gls*{HOP} can be realized using \gls*{MRM} based photonic integrated circuits. Each \gls*{MRM} corresponds to one hybrid photonic multiplication core, as shown in Fig. \ref{fig:principle}a. The \gls*{MRM} modulates a laser source of a particular wavelength. The input of the multiplier $d$ is loaded as binary words through the high-speed ports, while the weight $d$ from an N-bit \gls*{ADC} is loaded using a microheater-based modulation bias. The result is a weighted optical signal $y$, which is photodetected and sent to an \gls*{ADC} for decoding the information. The examples shown in Fig. \ref{fig:principle} with integer inputs and weights are for illustrative purposes. The multiplier can also handle normalized decimals, where the binary words represent only the decimals. Positive and negative weights are realized by biasing the \gls*{MRM} at the rising and falling slope of the transmission curve (Supplementary Figure 3). The relationship between the weight and the heater bias is measured and shown in Fig. \ref{fig:principle}b. The measurement is used as a lookup table for the loading of the normalized weights. This microheater-based weight loading supports refreshing at tens of kilohertz with thermal-optic effect. However, the refresh rate of the weight vector can be much quicker if the \gls*{HOP} was implemented using independent input and weight loading devices, for example, based on a Mach-Zehnder modulator and an \gls*{MRM} array\cite{bai23NC}. The \gls*{HOP} can therefore enable the possibility for in-situ training.

The implementation of the hybrid photonic \gls*{MVM} core is shown in Fig. \ref{fig:principle}c. The example illustrated here is a single 3$\times$3 convolution operator, but multiple convolution operators can be simultaneously integrated into a \gls*{PIC} to scale up for multiple convolution operations or complete matrix multiplication. A multi-wavelength light source is used with $L$ wavelengths for the convolution with an operator of $L$ elements. The \gls*{MRM} array accomplishes the element-wise multiplication of the input vector $(d_{11}, d_{21}, ..., d_{L1})^T$ and a weight vector (convolution operator) $(w_{11}, w_{12}, ..., w_{1L})$. $i$ denotes indices of the current column of the input matrix $D$. The example shows the input with 8-bit precision. The modulation generates a \gls*{WDM} signal with weighted \gls*{OOK} signaling for each wavelength (denoted as $\lambda$). The results from the hybrid photonic multipliers are summed up by a photodiode, accomplishing the convolution (\gls*{MVM}). The photodetection results in a multilevel signal at the baseband (i.e., a \gls*{PAM} signal). Shown in Fig. \ref{fig:principle}d, the multilevel signal is then further processed digitally word by word (time duration denoted as $t_{word}$) by a multilevel-to-binary converter, including \gls*{PAM} decoding (i.e., \gls*{ADC}), and a shift-add operation. The multilevel-to-binary conversion process converts the final result $y_n$ into the form of binary words (Details are discussed in Supplementary Figure 1).

\section*{Noise tolerance and computational robustness}

\begin{figure}[htbp]
    \centering
    \includegraphics[width=1\linewidth]{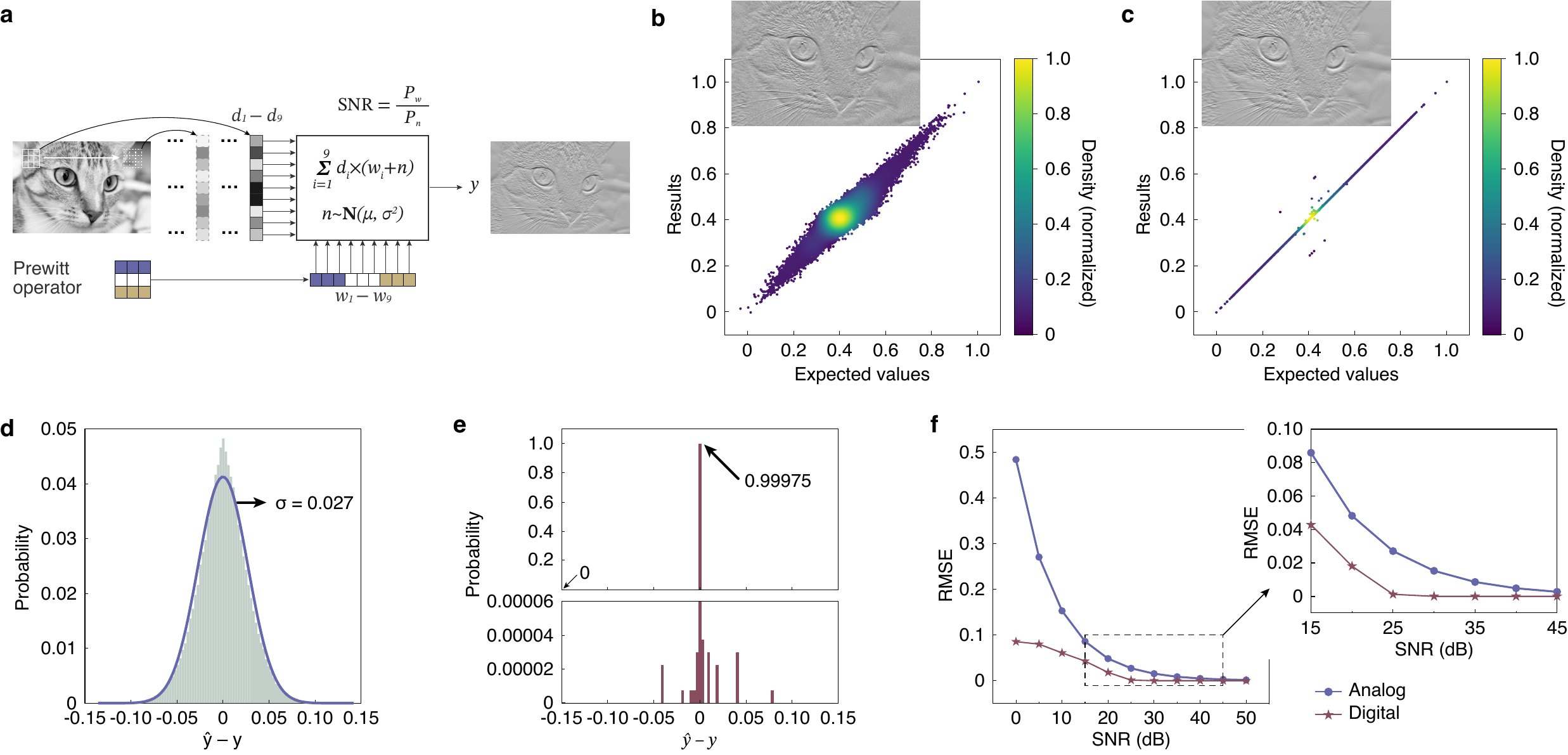}
    \caption{\textbf{Simulation setup and results.} \textbf{a,} Simulation setup. An image ``Chelsea'' from the scikit-image dataset\cite{van2014scikit} is convolved with a Prewitt operator (vertical edge detection). We explore the noise tolerance of both the analog and the hybrid optical computing systems by adding additive white Gaussian noise to the weights and examine the performance of the system by investigating the noise distribution of the outputs. \textbf{b,c,} Distribution of expected pixel values against the processed pixel values (both normalized) at an \gls*{SNR} of 25 dB, for the analog and hybrid computing systems, respectively. Insets show the corresponding processed images. Noisy pixels can be clearly observed in the image processed using analog computing. \textbf{d,e,} Noise distribution of the analog and hybrid computing systems, respectively, at an \gls*{SNR} of 25 dB. Analog computing reveals a Gaussian noise distribution with a standard deviation of 0.027, corresponding to a calculation precision of 3.6 bits. The \gls*{HOP} shows a greatly improved noise distribution thanks to the introduction of logic levels and decisions based on thresholding. \textbf{f,} performance of the analog and hybrid computing schemes in terms of \gls*{RMSE} with different \glspl*{SNR}.}
    \label{fig:simulations}
\end{figure}

\noindent We explore the noise tolerance of the proposed \gls*{HOP} using numerical simulations with a comparison to the analog computing scheme. Figure \ref{fig:simulations}a illustrates the simulation setup. An image ``Chelsea'' from the scikit-image dataset\cite{van2014scikit} is processed using the $3\times3$ Prewitt convolution operator for horizontal edge detection. We apply additive white Gaussian noise to the computing systems, i.e., the weights, with a given \gls*{SNR}. The image has a size of $300\times451$ pixels. The pixel values are normalized by feature scaling and reshaped into $3\times3$ data sequences, with the current vector denoted as $d_1-d_9$. The vector is multiplied with the Prewitt operator and the 9 results are summed up to form the pixel result of the processed image. For the analog scheme, the vector is encoded only on the amplitude of the signals. For the \gls*{HOP} scheme, the vector is encoded using binary words with 8-bit precision, representing a grey image with 8-bit color depth or 256 levels. The signal processing for the \gls*{HOP} scheme is done following procedures in Fig.\ref{fig:principle}c,d. Finally, the resulting pixels are reassembled to present the processed image, which has a size of $298\times449$, due to the lack of padding for the boundary of the original image. Figure \ref{fig:simulations}b,c give the distributions of the expected pixel values $y$ against the processed ones $\hat{y}$ at a \gls*{SNR} of 25 dB, for the analog and the \gls*{HOP} schemes, respectively. Insets show the processed and reconstructed images, where noisy pixels can be observed in the analog scheme, indicating a worse noise tolerance. 

Figure \ref{fig:simulations}d,e show the noise distribution (i.e., $\hat{y}-y$) of the output signals from the analog and the \gls*{HOP} schemes respectively. Results from the analog computing scheme reveal a Gaussian noise distribution with a standard deviation of 0.027. The corresponding calculation precision is 3.6 bits, calculated from the 3$\sigma$ value (99.7\% confidence interval) of the Gaussian-shaped noise distribution\cite{feldmann21Nature}. The noise distribution from the results of the \gls*{HOP} scheme is, however, fundamentally different from the analog computing scheme due to the introduction of logic levels and decisions. The noise in the results of the \gls*{HOP} scheme does not comply with a Gaussian distribution. Therefore, we borrowed the performance metrics from digital communication, i.e., error rate at a certain achievable \gls*{SNR} to characterize the \gls*{HOP} system. The \gls*{PER} is $2.5\times10^{-4}$ at an \gls*{SNR} of 25 dB, realizing a calculation precision of 8 bits. 

Figure \ref{fig:simulations}f gives the simulation results of the \gls*{RMSE} against different \glspl*{SNR} for both computing schemes. At an \gls*{SNR} of 25 dB, the \gls*{RMSE} can be reduced from $2.4\times10^{-2}$ using the analog computing scheme to $1.2\times10^{-3}$ using the \gls*{HOP} scheme. In short, the \gls*{HOP} scheme outperforms its analog counterpart in noise resilience, particularly within lower \gls*{SNR} regimes, suggesting superior scalability potential for optical computing applications. This resilience against noise is important, considering that noise presents a significant barrier to the scalability and practical deployment of \glspl*{ONN}.

\section*{Experimental setup}

\begin{figure}[htbp]
    \centering
    \includegraphics[width=0.821\linewidth]{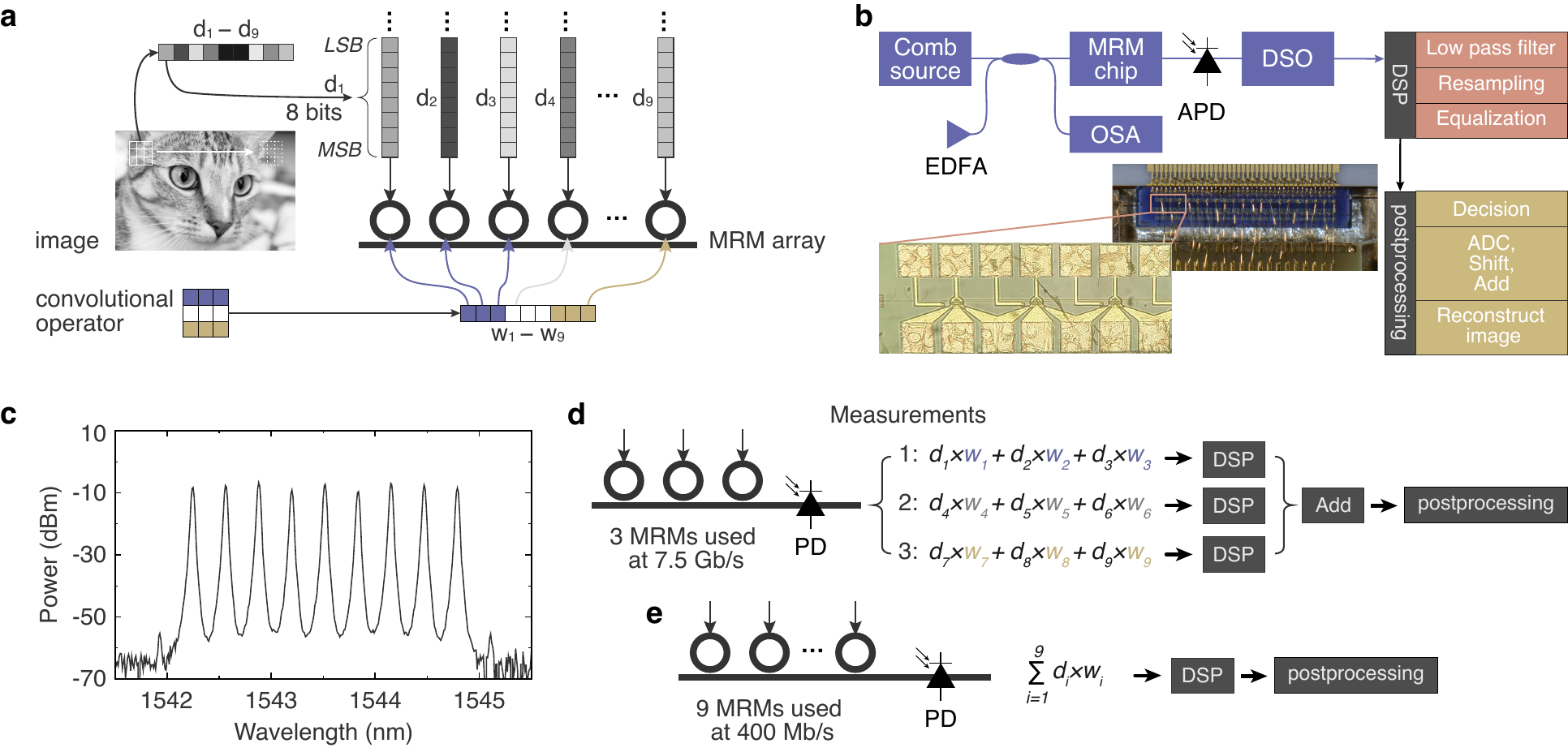}
    \caption{\textbf{Experimental setup.} \textbf{a,} Data loading scheme. The inputs of the \gls*{HOP} are the pixel values of the ``Chelsea'' image, used to perform the convolution operation. The inputs are loaded to the high-speed modulation ports of a set of 9 \glspl*{MRM} in binary words. The convolution operator is reshaped into a $1\times9$ vector and applied to the same set of \glspl*{MRM} using microheater-based modulation biases. \textbf{b,} Measurement setup, including \gls*{DSP} and signal postprocessing. The \gls*{HOP} consists of a packaged \gls*{PIC} chip containing 20 cascaded \glspl*{MRM}, and an external \gls*{PD}. Insets give the picture of the packaged chip and the microscopic image of the cascaded \glspl*{MRM}. \textbf{c,} The optical spectrum of the optical frequency comb source. 9 flattened comb lines are generated and fed into the \gls*{MRM} chip. \textbf{d,} Detailed operation condition and signal flow for the \acrshort*{HDIP} task. \textbf{e,} Detailed operation condition and signal flow for the \acrshort*{HWDR} task.}
    \label{fig:setup}
\end{figure}

\noindent We demonstrate the \gls*{HOP} in a proof-of-concept experiment. Figure \ref{fig:setup}a shows the data loading scheme. The inputs of the \gls*{HOP} come from the pixel values of the ``Chelsea'' image. The pixel values are normalized and reshaped into $3\times3$ data sequences, with the current vector denoted as $d_1-d_9$. The inputs are loaded to the high-speed modulation ports of an \gls*{MRM} array in the form of binary words and the weights from the convolution operator are applied to the corresponding \glspl*{MRM} using microheater-based modulation biases. Figure \ref{fig:setup}b shows the measurement setup. The multi-wavelength light source comes from a flattened \gls*{OFC} containing 9 wavelengths spaced at 80 GHz with an overall optical power of 7.0 dBm. The insets of \ref{fig:setup}b show the pictures of the packaged \gls*{PIC} chip and the microscopic image of the cascaded \glspl*{MRM}. The optical spectrum of the comb source is shown in Fig. \ref{fig:setup}c, with a central wavelength located around 1543.5 nm. An \gls*{EDFA} with a 10-dB optical coupler and an \gls*{OSA} are inserted after the comb source for noise loading and \gls*{OSNR} evaluations. The comb source is coupled into the \gls*{PIC} chip of cascaded \glspl*{MRM}\cite{kong20JLT,dong16OFC}. Each \gls*{MRM} modulates the corresponding comb line with the inputs and weights. The \gls*{WDM} signal generated from the \gls*{MRM} photonic chip is coupled into an \gls*{APD} with a launch power of $-$12.0 dBm. The photodetected baseband multilevel signal is sampled by a \gls*{DSO} and is further processed by a simple \gls*{DSP} chain\cite{kong21OFC}, including low-pass filtering, resampling, and most importantly equalization. Performance of the \gls*{HOP} is evaluated after the postprocessing procedure, including decision, multilevel-to-binary conversion (Fig. \ref{fig:principle}d), as well as image reconstruction. 

To assess the performance of the \gls*{HOP}, we undertook two tasks, each designed to probe specific functional aspects of the system. The \gls*{HDIP} task is designed to evaluate the calculation precision. The \gls*{HWDR} task is to evaluate the \gls*{HOP} in an inference task by a \gls*{CNN}. Depicted in Fig. \ref{fig:setup}d, the \gls*{HDIP} task is performed with each of the \glspl*{MRM} working at 7.5 Gb/s. The $3\times3$ convolution operator is disassembled into 3 sets and implemented through 3 measurements, due to our limited number of high-speed electrical signal channels. The results from the 3 measurements are added up together after \gls*{DSP} (before decision), and post-processing is done afterward for performance evaluation. Figure \ref{fig:setup}e shows the operation condition and signal flow for the \gls*{HWDR} task. We use an \gls*{FPGA} for simultaneously loading 9 microring modulators at a speed of 400 Mb/s to compute the entire convolution at once.

\section*{High-definition image processing and handwritten digit recognition}

\begin{figure}[htbp]
    \centering
    \includegraphics[width=0.9515\linewidth]{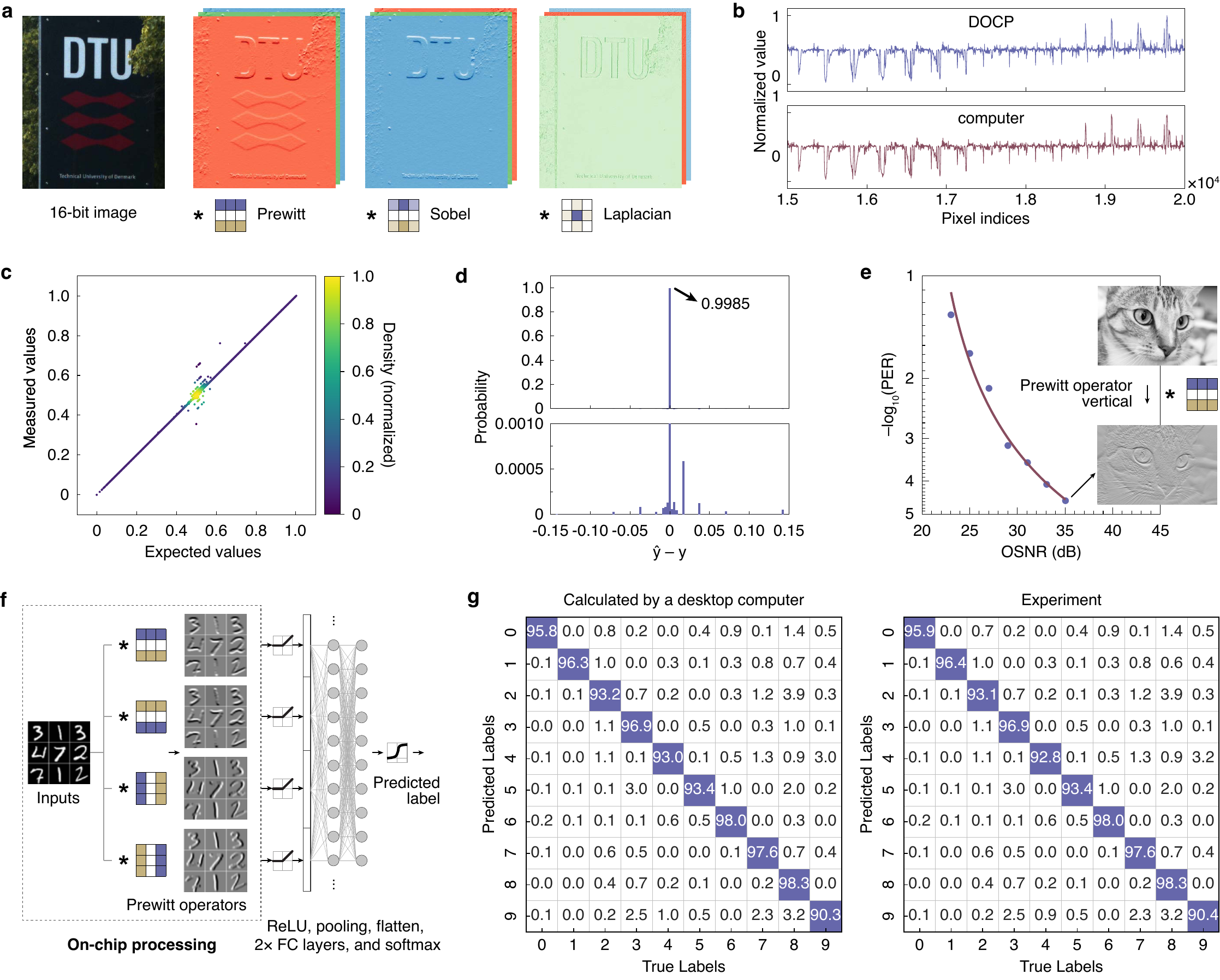}
    \caption{\textbf{Experiment results.} \textbf{a,} The original 16-bit image and the processed image channels using the Prewitt vertical, Sobel vertical and Laplacian operators. \textbf{b,} A section of the processed sequences of pixel values. Up: processed by the \gls*{HOP}; Down: processed by a desktop computer. \textbf{c,} Distribution of expected pixel values against the processed pixel values (both normalized). \textbf{d,} Noise distribution and calculation accuracy. \textbf{e,} Pixel error rate against light source \gls*{OSNR}. The measurements are performed on an 8-bit image processed with the Prewitt vertical operator. \textbf{f,} Layer structure of the \gls*{CNN} to perform the \gls*{HWDR} task using the \gls*{MNIST} database. The convolutional layer of the \gls*{CNN} is implemented using the \gls*{HOP} and the rest of the network is performed offline by a desktop computer. \textbf{g,} Confusion matrices for the prediction results, calculated by a desktop computer and the \gls*{HOP}.}
    \label{fig:results}
\end{figure}

\noindent Figure \ref{fig:results}a shows the original image and the processed red (R), Green (G), and Blue (B) channels. The image is taken from a mobile camera using uncompressed raw format. It is then converted to an image with a depth of 16 bits. The \gls*{HOP} processes the convolution using inputs with 16-bit binary words and applies the Prewitt vertical, Sobel vertical edge detection and Laplacian operators. The processed color channels remain in 16-bit depth and show a high-quality appearance indicating a high degree of noise resilience. This can be further proved by looking into the sequences of the pixel values with a comparison to the results calculated by a desktop computer, shown in Fig. \ref{fig:results}b. Figure \ref{fig:results}c shows the distribution of the expected pixel values $y$ against the processed ones $\hat{y}$. Figure \ref{fig:results}d derives the noise distribution (i.e., $\hat{y}-y$) of the results by the \gls*{HOP}. The corresponding measured \gls*{SNR} is 18.2 dB, and the \gls*{PER} is 1.8$\times\text{10}^{-3}$. We further tested the noise tolerance of the \gls*{HOP} system by introducing optical noise to the multi-wavelength light source. Figure \ref{fig:results}e gives the \gls*{PER} performance under tested \glspl*{OSNR} of the light source, for an 8-bit image processing with the Prewitt vertical operator. A \gls*{PER} of 6.0$\times\text{10}^{-5}$ is achieved at an \gls*{OSNR} of 35.0 dB. These results demonstrate that the \gls*{HOP} can handle the convolution task with very high precision, and the concept of digital-analog hybrid optical computing architecture works with a high tolerance to noise. More image results processed by different convolutional operators through \gls*{HOP} are demonstrated in Supplementary Figure 4.

Figure \ref{fig:results}f presents the layer structure of the \gls*{CNN} used for the \gls*{HWDR} task, utilizing the \gls*{MNIST} database\cite{MNIST}. 10,000 images are processed and classified based on one convolutional layer, rectified linear unit, pooling, flatten, as well as two fully connected layers of 100 and 10 neurons. We replaced the calculation of the convolution layer using our \gls*{HOP}, performing a full edge detection using four Prewitt operators. Figure \ref{fig:results}g illustrates the confusion matrices for the prediction results, calculated by a desktop computer and the \gls*{HOP}. We observed the same overall accuracy for the predictions compared with a desktop computer. We have also calculated the \glspl*{RMSE} of the 10,000 processed images compared with the true results calculated by a desktop computer. The results give a mean \gls*{RMSE} of $5.4\times10^{-3}$ and a standard deviation of $7.5\times10^{-3}$. And the overall \gls*{PER} for the 10,000 images is $2.7\times10^{-3}$. The results confirm that the \gls*{HOP} achieves a high level of calculation precision.

\section*{Discussion and outlook}
\noindent Optical computing has evolved from digital optical logic gates for general-purpose computing\cite{bogoni10OL} to domain-specific analog computing (for the physical implementation of neural networks\cite{shen17NP,xu21Nature}). Our domain-specific digital-analog hybrid optical computing architecture is conceptually different from logic-gates-based optical computing. Instead of pursuing general-purpose digital optical computing, we focus on domain-specified computing exclusively for matrix multiplications in neural networks. The implementation of matrix multiplication is based on binary modulations and linear signal processing, avoiding the nonlinear processes that are usually less efficient. Lastly, instead of pursuing an all-optical solution, the \gls*{HOP} combines the best of photonics for matrix multiplication and electronics for logic-level restoration. The \gls*{HOP} is distinctive from analog optical computing schemes for neural networks. Our results demonstrate the feasibility of overcoming the inherent challenges of analog optical computing through the digitization of optical signals. Given the benefits of better noise tolerance, the \gls*{HOP} has the potential to solve the obstacles of calculation precision, compatibility with microelectronics, and scaling of the \glspl*{ONN} (see Supplementary Table 3). Our findings here using digital optical inputs could potentially be applied to a wide range of \gls*{ONN} architectures, including the \gls*{MZI}-based coherent scheme\cite{shen17NP} and diffraction-based schemes\cite{luo22LSA,luo19LSA}.

\section*{Methods}
\subsection*{The multi-wavelength light source}

The multi-wavelength light source used in the experiment is an electro-optical frequency comb\cite{parriaux20AOP,kong15IPC} implemented using a continuous-wave laser and optical phase modulation. An external cavity laser centered around 1543.5 nm with an output power of 10.5 dBm is launched into a phase modulator. The phase modulator is driven by a 40 GHz signal coming from a \gls*{RF} synthesizer followed by a power amplifier. A 40-GHz spaced optical frequency comb is generated and fed into a \gls*{WSS} after amplification. It is line-by-line filtered and equalized by the \gls*{WSS}, generating a flattened optical frequency comb with 9 comb lines and an amplitude variation of 1 dB. The amplitude variation is not calibrated nor compensated for the computing system due to the noise tolerance of the \gls*{HOP}. In principle, compensating for residual system impairments, like the amplitude variation within the optical frequency comb, could further enhance the \gls*{HOP}'s performance.

\subsection*{The microring modulator array chip}

The \gls*{PIC} chip contains 20 cascaded \glspl*{MRM}, with a spacing of 250 $\mu$m between adjacent \glspl*{MRM}. Each \gls*{MRM} has a ring radius of 7.5 $\mu$m, thus a free spectral range around 13.1 nm. A microheater is sitting on top of each \gls*{MRM} for the alignment of the wavelength channels and the control of the modulation biases. The \glspl*{MRM} are based on a reverse-biased P-N junction in the middle of the microring waveguide \cite{dong16OFC, dong10OEMRM}. The \gls*{PIC} chip is fabricated on a standard silicon-on-insulator wafer with a top silicon thickness of 220 nm. It is packaged with two edge coupling fibers, a high-speed \gls*{PCB} supporting 20 high-speed transmission lines, and a second \gls*{PCB} for the control of the microheaters. The electro-optical modulation bandwidth of the packaged modulators is measured to be around 15 GHz. 

\subsection*{The signal source and digital signal processing}

In the \gls*{HDIP} task, an \gls*{AWG} (Keysight M8195A) is used. Limited by the available channels, 3 measurements are consecutively carried out. Each measurement uses 3 channels from the \gls*{AWG} at a speed of 7.5 Gb/s and a sampling rate of 15 GSa/s. The speed and the sampling rate are limited due to the use of external memory to load the entire image data, thus a mandatory clock division of four. In the \gls*{HWDR} task, an \gls*{FPGA} board (Xilinx ZCU104) is used to generate the 9 inputs to the \gls*{HOP}. The data rate for each data channel is set to 400 Mb/s due to the limitation of the direct memory access module. 

The weighted \gls*{WDM} signal is detected by an \gls*{APD} with a 3-dB bandwidth of 10 GHz. The photodetected baseband multilevel signal is sampled by a \gls*{DSO} (Agilent DSA-X 93304Q) with a sampling rate of 80 GSa/s and an analog bandwidth of 33 GHz. The samples are processed offline using \gls*{DSP} algorithms. We use a T/2-spaced linear feedforward equalizer with a filter length of 51 to compensate for the linear impairments from the devices used in the experiment, including the \gls*{RF} components, the high-speed \gls*{PCB} board, the \gls*{PIC} chip, as well as the \gls*{APD}. The coefficients of the equalizer are obtained through training based on the least-mean-square algorithm.

\section*{Data Availability}
Source data are provided with this paper. The measurement data generated in this study and processing scripts have also been deposited in \url{https://doi.org/10.5281/zenodo.10026199}. 

\section*{Code Availability}
The algorithms used for the data loading, reconstructions, and digital signal processing are standard and are outlined in detail in the Methods. Python scripts can be provided by the corresponding authors upon reasonable request.


\begin{thebibliography}{10}
\expandafter\ifx\csname url\endcsname\relax
  \def\url#1{\texttt{#1}}\fi
\expandafter\ifx\csname urlprefix\endcsname\relax\def\urlprefix{URL }\fi
\providecommand{\bibinfo}[2]{#2}
\providecommand{\eprint}[2][]{\url{#2}}

\bibitem{wu2023brief}
\bibinfo{author}{Wu, T.} \emph{et~al.}
\newblock \bibinfo{title}{A brief overview of chatgpt: The history, status quo and potential future development}.
\newblock \emph{\bibinfo{journal}{IEEE/CAA Journal of Automatica Sinica}} \textbf{\bibinfo{volume}{10}}, \bibinfo{pages}{1122--1136} (\bibinfo{year}{2023}).

\bibitem{Amodei18}
\bibinfo{author}{Amodei, D.} \& \bibinfo{author}{Hernandez, D.}
\newblock \bibinfo{title}{Review of ai and compute}.
\newblock \bibinfo{howpublished}{\url{https://openai.com/blog/ai-and-compute/}} (\bibinfo{year}{2018}).

\bibitem{jouppi17}
\bibinfo{author}{Jouppi, N.~P.} \emph{et~al.}
\newblock \bibinfo{title}{In-datacenter performance analysis of a tensor processing unit}.
\newblock In \emph{\bibinfo{booktitle}{Proceedings of the 44th annual international symposium on computer architecture}}, \bibinfo{pages}{1--12} (\bibinfo{year}{2017}).

\bibitem{merolla14}
\bibinfo{author}{Merolla, P.~A.} \emph{et~al.}
\newblock \bibinfo{title}{A million spiking-neuron integrated circuit with a scalable communication network and interface}.
\newblock \emph{\bibinfo{journal}{Science}} \textbf{\bibinfo{volume}{345}}, \bibinfo{pages}{668--673} (\bibinfo{year}{2014}).

\bibitem{mooreslaw}
\bibinfo{author}{Theis, T.~N.} \& \bibinfo{author}{Wong, H.-S.~P.}
\newblock \bibinfo{title}{The end of moore's law: A new beginning for information technology}.
\newblock \emph{\bibinfo{journal}{Computing in Science Engineering}} \textbf{\bibinfo{volume}{19}}, \bibinfo{pages}{41--50} (\bibinfo{year}{2017}).

\bibitem{rupp21}
\bibinfo{author}{Rupp, K.} \emph{et~al.}
\newblock \bibinfo{title}{42 years of microprocessor trend data}.
\newblock \emph{\bibinfo{journal}{https://www.karlrupp.net/ 2018/02/42-years-of-microprocessor-trend-data/ [Online]}}  (\bibinfo{year}{2018}).

\bibitem{khan18}
\bibinfo{author}{Khan, H.~N.}, \bibinfo{author}{Hounshell, D.~A.} \& \bibinfo{author}{Fuchs, E.~R.}
\newblock \bibinfo{title}{Science and research policy at the end of moore’s law}.
\newblock \emph{\bibinfo{journal}{Nature Electronics}} \textbf{\bibinfo{volume}{1}}, \bibinfo{pages}{14--21} (\bibinfo{year}{2018}).

\bibitem{caulfield10NP}
\bibinfo{author}{Caulfield, H.~J.} \& \bibinfo{author}{Dolev, S.}
\newblock \bibinfo{title}{Why future supercomputing requires optics}.
\newblock \emph{\bibinfo{journal}{Nature Photonics}} \textbf{\bibinfo{volume}{4}}, \bibinfo{pages}{261--263} (\bibinfo{year}{2010}).

\bibitem{miller15Optica}
\bibinfo{author}{Miller, D. A.~B.}
\newblock \bibinfo{title}{Perfect optics with imperfect components}.
\newblock \emph{\bibinfo{journal}{Optica}} \textbf{\bibinfo{volume}{2}}, \bibinfo{pages}{747--750} (\bibinfo{year}{2015}).

\bibitem{peng18JSTQE}
\bibinfo{author}{Peng, H.-T.}, \bibinfo{author}{Nahmias, M.~A.}, \bibinfo{author}{de~Lima, T.~F.}, \bibinfo{author}{Tait, A.~N.} \& \bibinfo{author}{Shastri, B.~J.}
\newblock \bibinfo{title}{Neuromorphic photonic integrated circuits}.
\newblock \emph{\bibinfo{journal}{IEEE Journal of Selected Topics in Quantum Electronics}} \textbf{\bibinfo{volume}{24}}, \bibinfo{pages}{1--15} (\bibinfo{year}{2018}).

\bibitem{shen17NP}
\bibinfo{author}{Shen, Y.} \emph{et~al.}
\newblock \bibinfo{title}{{Deep learning with coherent nanophotonic circuits}}.
\newblock \emph{\bibinfo{journal}{Nature Photonics}} \textbf{\bibinfo{volume}{11}}, \bibinfo{pages}{441--446} (\bibinfo{year}{2017}).

\bibitem{feldmann21Nature}
\bibinfo{author}{Feldmann, J.} \emph{et~al.}
\newblock \bibinfo{title}{{Parallel convolutional processing using an integrated photonic tensor core}}.
\newblock \emph{\bibinfo{journal}{Nature}} \textbf{\bibinfo{volume}{589}}, \bibinfo{pages}{52--58} (\bibinfo{year}{2021}).

\bibitem{nahmias19}
\bibinfo{author}{Nahmias, M.~A.} \emph{et~al.}
\newblock \bibinfo{title}{Photonic multiply-accumulate operations for neural networks}.
\newblock \emph{\bibinfo{journal}{IEEE Journal of Selected Topics in Quantum Electronics}} \textbf{\bibinfo{volume}{26}}, \bibinfo{pages}{1--18} (\bibinfo{year}{2019}).

\bibitem{tait16}
\bibinfo{author}{Tait, A.~N.}, \bibinfo{author}{De~Lima, T.~F.}, \bibinfo{author}{Nahmias, M.~A.}, \bibinfo{author}{Shastri, B.~J.} \& \bibinfo{author}{Prucnal, P.~R.}
\newblock \bibinfo{title}{Multi-channel control for microring weight banks}.
\newblock \emph{\bibinfo{journal}{Optics Express}} \textbf{\bibinfo{volume}{24}}, \bibinfo{pages}{8895--8906} (\bibinfo{year}{2016}).

\bibitem{tait17}
\bibinfo{author}{Tait, A.~N.} \emph{et~al.}
\newblock \bibinfo{title}{Neuromorphic photonic networks using silicon photonic weight banks}.
\newblock \emph{\bibinfo{journal}{Scientific reports}} \textbf{\bibinfo{volume}{7}}, \bibinfo{pages}{1--10} (\bibinfo{year}{2017}).

\bibitem{zhang22Optica}
\bibinfo{author}{Zhang, W.} \emph{et~al.}
\newblock \bibinfo{title}{Silicon microring synapses enable photonic deep learning beyond 9-bit precision}.
\newblock \emph{\bibinfo{journal}{Optica}} \textbf{\bibinfo{volume}{9}}, \bibinfo{pages}{579--584} (\bibinfo{year}{2022}).

\bibitem{bai23NC}
\bibinfo{author}{Bai, B.} \emph{et~al.}
\newblock \bibinfo{title}{Microcomb-based integrated photonic processing unit}.
\newblock \emph{\bibinfo{journal}{Nature Communications}} \textbf{\bibinfo{volume}{14}}, \bibinfo{pages}{Article number: 66} (\bibinfo{year}{2023}).

\bibitem{gupta15}
\bibinfo{author}{Gupta, S.}, \bibinfo{author}{Agrawal, A.}, \bibinfo{author}{Gopalakrishnan, K.} \& \bibinfo{author}{Narayanan, P.}
\newblock \bibinfo{title}{Deep learning with limited numerical precision}.
\newblock In \emph{\bibinfo{booktitle}{International conference on machine learning}}, \bibinfo{pages}{1737--1746} (\bibinfo{organization}{PMLR}, \bibinfo{year}{2015}).

\bibitem{adc_survey}
\bibinfo{author}{Murmann, B.}
\newblock \bibinfo{title}{{ADC Performance Survey 1997-2023}}.
\newblock \bibinfo{note}{[Online]. Available: \url{https://github.com/bmurmann/ADC-survey}}.

\bibitem{van2014scikit}
\bibinfo{author}{Van~der Walt, S.} \emph{et~al.}
\newblock \bibinfo{title}{scikit-image: image processing in python}.
\newblock \emph{\bibinfo{journal}{PeerJ}} \textbf{\bibinfo{volume}{2}}, \bibinfo{pages}{e453} (\bibinfo{year}{2014}).

\bibitem{kong20JLT}
\bibinfo{author}{Kong, D.} \emph{et~al.}
\newblock \bibinfo{title}{{Intra-datacenter interconnects with a serialized silicon optical frequency comb modulator}}.
\newblock \emph{\bibinfo{journal}{Journal of Lightwave Technology}} \textbf{\bibinfo{volume}{38}}, \bibinfo{pages}{4677--4682} (\bibinfo{year}{2020}).

\bibitem{dong16OFC}
\bibinfo{author}{Dong, P.}, \bibinfo{author}{Lee, J.}, \bibinfo{author}{Kim, K.}, \bibinfo{author}{Chen, Y.-K.} \& \bibinfo{author}{Gui, C.}
\newblock \bibinfo{title}{Ten-channel discrete multi-tone modulation using silicon microring modulator array}.
\newblock In \emph{\bibinfo{booktitle}{2016 Optical Fiber Communications Conference and Exhibition (OFC)}}, \bibinfo{pages}{1--3} (\bibinfo{year}{2016}).

\bibitem{kong21OFC}
\bibinfo{author}{Kong, D.} \emph{et~al.}
\newblock \bibinfo{title}{{100 Gbit/s PAM-16 Transmission in the 2-$\mu$m Band over a 1.15-km Hollow-Core Fiber}}.
\newblock In \emph{\bibinfo{booktitle}{Optical Fiber Communications Conference (OFC) 2021}}, \bibinfo{pages}{Th4E.6} (\bibinfo{organization}{IEEE}, \bibinfo{year}{2021}).

\bibitem{MNIST}
\bibinfo{author}{LeCun, Y.}, \bibinfo{author}{Cortes, C.} \& \bibinfo{author}{Burges, C.~J.}
\newblock \bibinfo{title}{{The MNIST database of handwritten digits}}.
\newblock \bibinfo{note}{[Online]. Available: \url{http://yann.lecun.com/exdb/mnist/}}.

\bibitem{bogoni10OL}
\bibinfo{author}{Bogoni, A.}, \bibinfo{author}{Wu, X.}, \bibinfo{author}{Bakhtiari, Z.}, \bibinfo{author}{Nuccio, S.} \& \bibinfo{author}{Willner, A.~E.}
\newblock \bibinfo{title}{640 gbits/s photonic logic gates}.
\newblock \emph{\bibinfo{journal}{Opt. Lett.}} \textbf{\bibinfo{volume}{35}}, \bibinfo{pages}{3955--3957} (\bibinfo{year}{2010}).

\bibitem{xu21Nature}
\bibinfo{author}{Xingyuan, X.} \emph{et~al.}
\newblock \bibinfo{title}{{11 TOPS photonic convolutional accelerator for optical neural networks}}.
\newblock \emph{\bibinfo{journal}{Nature}} \textbf{\bibinfo{volume}{589}}, \bibinfo{pages}{44--51} (\bibinfo{year}{2021}).

\bibitem{luo22LSA}
\bibinfo{author}{Luo, X.} \emph{et~al.}
\newblock \bibinfo{title}{Metasurface-enabled on-chip multiplexed diffractive neural networks in the visible}.
\newblock \emph{\bibinfo{journal}{Light: Science \& Applications}} \textbf{\bibinfo{volume}{11}}, \bibinfo{pages}{Article number: 158} (\bibinfo{year}{2022}).

\bibitem{luo19LSA}
\bibinfo{author}{Luo, Y.} \emph{et~al.}
\newblock \bibinfo{title}{Design of task-specific optical systems using broadband diffractive neural networks}.
\newblock \emph{\bibinfo{journal}{Light: Science \& Applications}} \textbf{\bibinfo{volume}{8}}, \bibinfo{pages}{Article number: 112} (\bibinfo{year}{2019}).

\bibitem{parriaux20AOP}
\bibinfo{author}{Parriaux, A.}, \bibinfo{author}{Hammani, K.} \& \bibinfo{author}{Millot, G.}
\newblock \bibinfo{title}{Electro-optic frequency combs}.
\newblock \emph{\bibinfo{journal}{Adv. Opt. Photon.}} \textbf{\bibinfo{volume}{12}}, \bibinfo{pages}{223--287} (\bibinfo{year}{2020}).

\bibitem{kong15IPC}
\bibinfo{author}{Kong, D.} \emph{et~al.}
\newblock \bibinfo{title}{{Cavity-less sub-picosecond pulse generation for the demultiplexing of a 640 Gbaud OTDM signal}}.
\newblock In \emph{\bibinfo{booktitle}{IEEE Photonics Conference (IPC) 2015}}, \bibinfo{pages}{WG1.2} (\bibinfo{organization}{IEEE}, \bibinfo{year}{2015}).

\bibitem{dong10OEMRM}
\bibinfo{author}{Dong, P.} \emph{et~al.}
\newblock \bibinfo{title}{Wavelength-tunable silicon microring modulator}.
\newblock \emph{\bibinfo{journal}{Opt. Express}} \textbf{\bibinfo{volume}{18}}, \bibinfo{pages}{10941--10946} (\bibinfo{year}{2010}).

\end{thebibliography}


\section*{Acknowledgements}
D.K. acknowledges the research grant (VIL53077) of the Young Investigator Program (DONN) from the VILLUM FONDEN. H.H. acknowledges the research grant (VIL34320) of the Synergy Program (Searchlight) from the VILLUM FONDEN. 

\section*{Author Contributions Statement}
X.M. and D.K. conceived the concept and the experiment; X.M. and Q.L. performed the simulations, supervised by D.K., I.C. and C.L.; D.K. and X.M. designed the experiment; X.M. and D.K. constructed the experiment setup, performed the experiment, and processed the data; P.D. designed the \gls*{PIC}; K.K. packaged the \gls*{PIC}; D.K. and X.M. characterized the \gls*{PIC}; X.M., D.K., I.C., C.L. and H.H. discussed the results; The manuscript was written by D.K. and X.M., and all authors contributed to the writing; D.K. and H.H. supervised the projects.

\section*{Competing Interests Statement}
The authors declare no competing interests. 

\section*{Figure Legends}
\noindent{\textbf{Fig. 1 $\mid$ Concept of the digital-analog hybrid photonic MVM core.} \textbf{a,} Digital signal is more robust to noise and crosstalk in a signal processing system. \Acrshort*{ONN} can be seen as a signal processing system where digital signals can potentially be applied for better calculation repeatability, precision, scalability, and compatibility with microelectronics compared to analog signals. \textbf{b,} The abstracted analog photonic \acrshort*{MVM} core utilizing analog signals for both input $d$ and weight $w$. \textbf{c,} The proposed digital-analog hybrid photonic \acrshort*{MVM} core utilizing digital signals for input $d$, with relaxed constraints for signal format converters.}
\newline

\noindent{\textbf{Fig. 2 $\mid$ A \acrshort*{MRM} based implementation of the digital-analog hybrid photonic \acrshort*{MVM} core.} \textbf{a,} The implementation of a single hybrid digital-analog photonic multiplication core using an \acrshort*{MRM}, where the input $d$ in the form of a digital optical signal is loaded through the high-speed port while the weight $w$ is loaded using microheater based modulation bias. \textbf{b,} The measured relationship between the normalized weight and the required heater voltage for the modulation bias, can be used as a lookup table to load the weight. \textbf{c,} The implementation and optical signal temporal evolution of the digital-analog hybrid photonic \acrshort*{MVM} core, including a multi-wavelength light source, an array of microring modulators, and a photodiode. \textbf{d,} Post-processing of the multilevel signal includes an \acrshort*{ADC} and a shift-add operation. Here the multilevel signal is converted to a binary signal and the final result can be recovered via shifts and adds.}
\newline

\noindent{\textbf{Fig. 3 $\mid$ Simulation setup and results.} \textbf{a,} Simulation setup. An image ``Chelsea'' from the scikit-image dataset\cite{van2014scikit} is convolved with a Prewitt operator (vertical edge detection). We explore the noise tolerance of both the analog and the hybrid optical computing systems by adding additive white Gaussian noise to the weights and examine the performance of the system by investigating the noise distribution of the outputs. \textbf{b,c,} Distribution of expected pixel values against the processed pixel values (both normalized) at an \gls*{SNR} of 25 dB, for the analog and hybrid computing systems, respectively. Insets show the corresponding processed images. Noisy pixels can be clearly observed in the image processed using analog computing. \textbf{d,e,} Noise distribution of the analog and hybrid computing systems, respectively, at an \gls*{SNR} of 25 dB. Analog computing reveals a Gaussian noise distribution with a standard deviation of 0.027, corresponding to a calculation precision of 3.6 bits. The \gls*{HOP} shows a greatly improved noise distribution thanks to the introduction of logic levels and decisions based on thresholding. \textbf{f,} performance of the analog and hybrid computing schemes in terms of \gls*{RMSE} with different \glspl*{SNR}.}
\newline

\noindent{\textbf{Fig. 4 $\mid$ Experimental setup.} \textbf{a,} Data loading scheme. The inputs of the \gls*{HOP} are the pixel values of the ``Chelsea'' image, used to perform the convolution operation. The inputs are loaded to the high-speed modulation ports of a set of 9 \glspl*{MRM} in binary words. The convolution operator is reshaped into a $1\times9$ vector and applied to the same set of \glspl*{MRM} using microheater-based modulation biases. \textbf{b,} Measurement setup, including \gls*{DSP} and signal postprocessing. The \gls*{HOP} consists of a packaged \gls*{PIC} chip containing 20 cascaded \glspl*{MRM}, and an external \gls*{PD}. Insets give the picture of the packaged chip and the microscopic image of the cascaded \glspl*{MRM}. \textbf{c,} The optical spectrum of the optical frequency comb source. 9 flattened comb lines are generated and fed into the \gls*{MRM} chip. \textbf{d,} Detailed operation condition and signal flow for the \acrshort*{HDIP} task. \textbf{e,} Detailed operation condition and signal flow for the \acrshort*{HWDR} task.}
\newline

\noindent{\textbf{Fig. 5 $\mid$ Experiment results.} \textbf{a,} The original 16-bit image and the processed image channels using the Prewitt vertical, Sobel vertical and Laplacian operators. \textbf{b,} A section of the processed sequences of pixel values. Up: processed by the \gls*{HOP}; Down: processed by a desktop computer. \textbf{c,} Distribution of expected pixel values against the processed pixel values (both normalized). \textbf{d,} Noise distribution and calculation accuracy. \textbf{e,} Pixel error rate against light source \gls*{OSNR}. The measurements are performed on an 8-bit image processed with the Prewitt vertical operator. \textbf{f,} Layer structure of the \gls*{CNN} to perform the \gls*{HWDR} task using the \gls*{MNIST} database. The convolutional layer of the \gls*{CNN} is implemented using the \gls*{HOP} and the rest of the network is performed offline by a desktop computer. \textbf{g,} Confusion matrices for the prediction results, calculated by a desktop computer and the \gls*{HOP}.}

\end{document}



\maketitle
\vspace{-40pt}
\section*{Corresponding authors}
Correspondence to: \\
Deming Kong (dmkon@dtu.dk), ORCID: 0000-0001-6552-4081\\
Hao Hu (huhao@dtu.dk), ORCID: 0000-0002-8859-0986


\section{Post-processing circuit for the \acrlong*{HOP}}
The post-processing converts the multi-level signal into the final calculation result and can be implemented using microelectronic circuits. We use an 8-bit wide calculated result as an example in the following discussion. The output of the \gls*{HOP} is a \gls*{PAM}-like signal. It is decoded and further converted into parallel 8-bit digital sequences by a high-speed \gls*{ADC}, as depicted in Supplementary Figure \ref{Post_processing}a, b. The high-speed digital sequences are then transformed into low-speed digital binaries using a deserializer. For instance, a single output sequence of the \gls*{ADC} with a clock speed of 7.5 GHz can be down-converted to $1/8$ speed through a deserializer, i.e., to a speed of 937.5 MHz, as shown in Supplementary Figure \ref{Post_processing}c. The deserializing process facilitates subsequent electrical circuits. The converted data are then routed to a logic circuit, as shown in Supplementary Figure \ref{Post_processing}d, which resolves the final calculated result. This circuitry works similarly to an 8-bit multiplier, including a number of half-adders and full-adders, but without AND gates preceding the adders. Although the circuit can introduce some gate latency, as illustrated in Supplementary Figure \ref{Post_processing}e, the whole post-processing can be accomplished in a single electrical clock cycle.

\begin{figure}[htbp]
    \centering
    \includegraphics[width=0.99\linewidth]{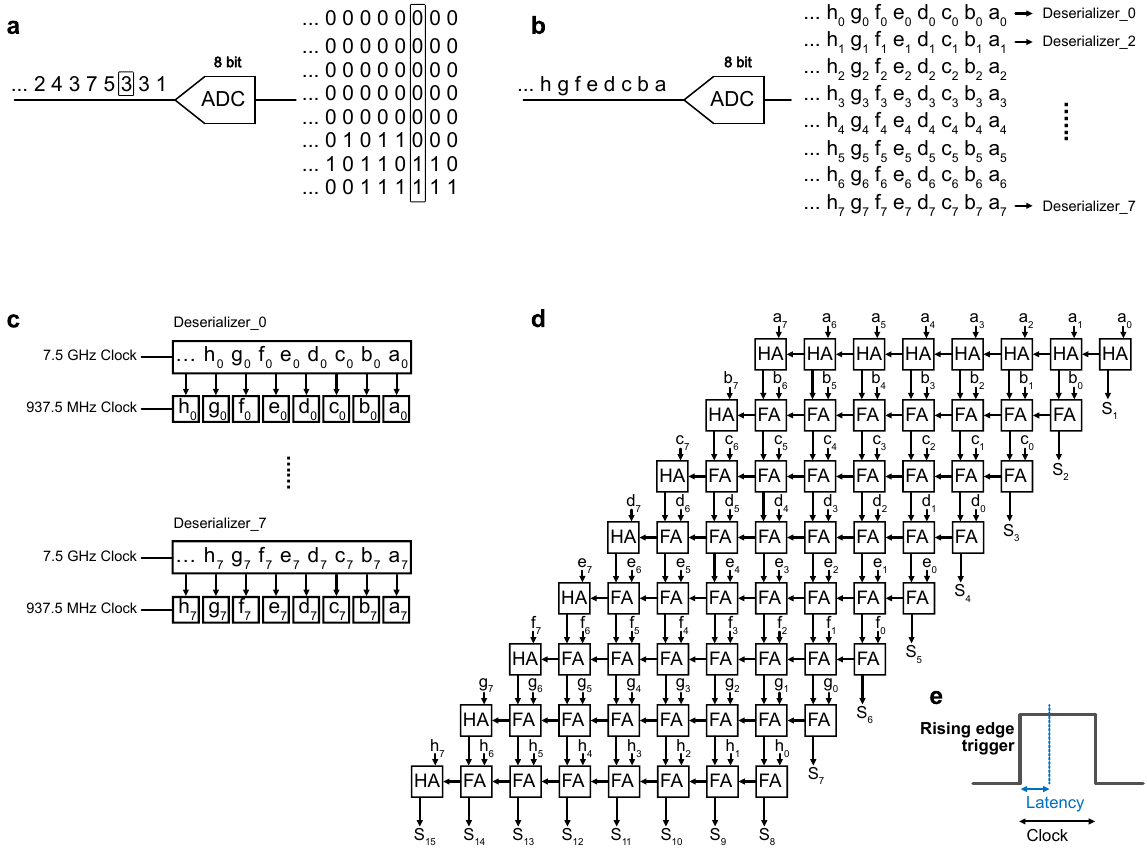}
    \caption{\textbf{Post-processing circuit for the \acrlong*{HOP}.} \textbf{a-b,} High-speed \gls*{ADC} converts the calculation result carried with the \gls*{PAM}-like signal into parallel digital sequences. \textbf{c,} Deserializers down-convert the high-speed signal, e.g., at a clock speed of 7.5 GHz, to a more manageable 937.5 MHz binaries. \textbf{d,} The final calculation result is resolved by a digital circuit composed of half-adders and full-adders. \textbf{e,} The processing circuit introduces some gate latency, but the whole process can be accomplished within a single electrical clock cycle.}
    \label{Post_processing}
\end{figure}

\section{Detailed comparison between the \acrlong*{HOP} and analog computing schemes}
We show theoretically the computing power of the \gls*{HOP} and compare it with analog optical computing schemes in the following discussions. As previously discussed in the paper (see Figure 1b and 1c), the output signals of both schemes are processed by the \glspl*{ADC} and the resolution (\gls*{ENOB}) requirement are as follows: 
\vspace{-10pt}

\begin{equation}\label{enob_analog}
\text{ENOB}_\text{Analog} = \text{log}_2\left[L\times\left(2^M-1\right)\times\left(2^N-1\right)\right]
\end{equation}

\vspace{-20pt}

\begin{equation}\label{enob_hybrid}
\text{ENOB}_\text{Hybrid} = \text{log}_{2}\left[L\times\left(2^N-1\right)\right]
\end{equation}

\noindent where M and N denote the resolution (bit widths) of input data 'd' and weight value 'w', respectively; L is the number of multiplications for the inner product of the input (multiplicand) vector and the weight (multiplier) vector, representing the scale of the convolutional operator and optical hardware. Given that the input in the hybrid scheme is binary data, 'M' equals 1. The required \gls*{ENOB} of the \gls*{ADC} increases with the scale of the weight vector and the bit width of each single weight in both schemes. Yet, the required \gls*{ENOB} for the analog scheme also increases with the bit width of the inputs. This means that, with the same hardware scale, the required \gls*{ENOB} of the analog scheme surpasses its counterpart in the \gls*{HOP} scheme when the bit width of the inputs is greater than 2.

However, it is known in practice\cite{adc_survey} that the maximum sampling rate of the \gls*{ADC} is inversely proportional to $2^{\text{ENOB}}$
\vspace{-10pt}

\begin{equation}\label{adc_limit}
F_s = \frac{c}{2^{\text{ENOB}}}
\end{equation}

\noindent where c is a constant, with a typical value of 5 TS/s-levels for cutting-edge \glspl*{ADC} implementations. Utilizing equations (\ref{enob_analog}) and (\ref{enob_hybrid}), the maximum sampling rate (i.e., the speed of operations) to ensure a certain calculation precision for both the analog and hybrid computing schemes can be calculated. We can see that the achievable sampling rate of the \gls*{ADC} is inversely proportional to the scale for a single convolutional operator.

\vspace{-10pt}

\begin{equation}\label{fs_analog}
F_{s}(\text{Analog}) = \frac{5}{L\times\left(2^M-1\right)\times\left(2^N-1\right)} \hspace{10pt}\text{TS/s}
\end{equation}

\vspace{-10pt}

\begin{equation}\label{fs_hybrid}
F_{s}(\text{Hybrid}) = \frac{5}{L\times\left(2^N-1\right)} \hspace{10pt}\text{TS/s}
\end{equation}

\noindent Then the computing power $\text{P}$ for a single operator in terms of \gls*{TOPS} for two schemes can be further calculated by multiplying the hardware scale (i.e., L) with the maximum achievable speed of the system (proportional to the sampling rate of the \glspl*{ADC}, i.e., $F_{s}$), shown as equations (\ref{tops_analog}) and (\ref{tops_hybrid}). 

\begin{equation}\label{tops_analog}
\text{P}_\text{Analog} \propto \frac{5}{\left(2^M-1\right)\times\left(2^N-1\right)} \hspace{10pt}\text{TOPS}
\end{equation}

\begin{equation}\label{tops_hybrid}
\text{P}_\text{Hybrid} \propto \frac{5}{p\times\left(2^N-1\right)} \hspace{10pt}\text{TOPS}
\end{equation}

\noindent where $p$ is the bit width (precision) of the binary inputs and the multi-level outputs.

For the analog computing scheme, we have assumed that the \glspl*{DAC} will not impose any speed or precision limit. Therefore, the computing speed is determined by the achievable operating speed of the \glspl*{ADC}. In practice, the performance of the analog computing scheme is also limited by the \glspl*{DAC}. This assumption here is used for simplicity reasons only, thus the calculated performance for the analog scheme is overestimated. 

For the hybrid computing scheme, no \glspl*{DAC} is used. But the input data is binary words with $p$-bit precision, thus $p$ times slower than the analog scheme, in theory. For example, if we assume 8-bit input data with a clock rate of 10 GHz for both the analog scheme using \glspl*{DAC} and the hybrid scheme using pattern generators (0s and 1s), then the input rate for binary words in the hybrid scheme is divided by 8, i.e., 1.25 GHz.

A major conclusion can be drawn from equations (\ref{tops_analog}) and (\ref{tops_hybrid}). The computing power, \gls*{TOPS}, does not scale with the size for a single convolutional operator. Along with the increase in the scale of the hardware (i.e., $L$), the requirement for \gls*{ENOB} is increased, leading to a reduction of the sampling rate of the \gls*{ADC}(equation (\ref{fs_analog}) and (\ref{fs_hybrid})). The hardware scale $L$ is eventually canceled out in the \gls*{TOPS}. This also indicates that the \glspl*{ADC} are the main bottlenecks of the computing systems, both for the analog scheme and the hybrid scheme. Note that this conclusion should not only apply to photonic computing schemes but also to microelectronics-based analog computing. 

To overcome this, one can implement multiple operators to increase the overall computing power for the whole computing system. In other words, multiple low-resolution \glspl*{ADC} can be utilized together with small-scale operators (resulting in fewer signal levels and thus possible higher \gls*{ADC} sampling rate), instead of a large \gls*{MVM} operator with a high-resolution \gls*{ADC}. The study of \gls*{ADC}\cite{adc_survey} demonstrates that the requirement for logic gate resources increases exponentially with the resolution of the \gls*{ADC}. Therefore, deploying multiple low-resolution \glspl*{ADC} consumes a similar amount of logic gate resources as a single high-precision \glspl*{ADC}. 

The above approach could increase the sampling rate of the \glspl*{ADC} without sacrificing the overall electrical hardware resource, thus achieving a higher computing power of the computing system. Therefore, the \gls*{HOP} is particularly well-suited for processing tasks that require a large number of small convolutional operators, such as the widely-known object detection model, \gls*{YOLO}\cite{redmon2018yolov3}. 

\begin{table}[ht]
\begin{threeparttable}
\centering
\begin{tabular}{|C{2cm}|C{1.2cm}|C{1.2cm}|C{1.2cm}|C{1.2cm}|C{1.2cm}|C{1.2cm}|C{1.2cm}|C{1.2cm}|}
\hline
\rowcolor{Gainsboro!100}
Input Data Precision & 1 bit &  2 bit & 3 bit& 4 bit &  5 bit & 6 bit &  7 bit & 8 bit \\ \hline
Required \gls*{ADC} \gls*{ENOB} & 3.3 bit  & 6.4 bit & 8.8 bit & 11 bit & 13.1 bit & 15.1 bit & 17.1 bit & 19.2 bit \\ \hline
\gls*{ADC} Achievable Speed & 500 GHz & 60.9 GHz  & 11.3 GHz & 2.5 GHz  & 578 MHz & 140 MHz  & 34.4 MHz & 8.5 MHz \\ \hline
\gls*{DAC} Speed & 40 GHz & 40 GHz  & 40 GHz & 40 GHz  & 40 GHz & 40 GHz  & 40 GHz & 40 GHz \\ \hline
System Speed\footnotemark[1] & 40  GHz & 40 GHz  & 11.3 GHz & 2.5 GHz  & 578 MHz & 140 MHz  & 34.4 MHz & 8.5 MHz \\ \hline
TOPS\footnotemark[2] & 3.6E-1 & 3.6E-1  & 1E-1 & 2.2E-2  & 5.2E-3 & 1.2E-3  & 3.1E-4
 & 8E-5 \\ \hline
\end{tabular}
\caption{TOPS performance of analog optical processor with various different calculation precision levels.}
\label{Analog_scheme}   

\begin{tablenotes}
    \item[1] The system speed is determined by the lesser of two speeds: the maximum achievable speed of the \gls*{ADC} and the input data speed through the \gls*{DAC}.
    \item[2] The convolution task utilizes a 3$\times$3 size kernel, and the weight value precision is 8 bit. The TOPS is calculated by multiplying the hardware scale (i.e., L = 9) with the speed of the system.
    
\end{tablenotes}
\end{threeparttable}
\end{table}

\begin{table}[ht]
\begin{threeparttable}
\centering
\begin{tabular}{|C{2cm}|C{1.2cm}|C{1.2cm}|C{1.2cm}|C{1.2cm}|C{1.2cm}|C{1.2cm}|C{1.2cm}|C{1.2cm}|}
\hline
\rowcolor{Gainsboro!100}
Input Data Precision & 1 bit &  2 bit & 3 bit& 4 bit &  5 bit & 6 bit &  7 bit & 8 bit \\ \hline
Required \gls*{ADC} \gls*{ENOB} & 3.3 bit  & 4.8 bit  & 6.0 bit & 7.1 bit & 8.1 bit & 9.1 bit & 10.2 bit & 11.2 bit \\ \hline
\gls*{ADC} Achievable Speed & 500 GHz & 178 GHz  & 78 GHz & 36.8 GHz  & 17.9 GHz & 8.8 GHz  & 4.4 GHz & 2.2 GHz \\ \hline
Digital I/O Speed & 40 GHz & 40 GHz  & 40 GHz & 40 GHz  & 40 GHz & 40 GHz  & 40 GHz & 40 GHz \\ \hline
System Speed\footnotemark[1] & 40 GHz & 40 GHz  & 40 GHz & 36.8 GHz  & 17.9 GHz & 8.8 GHz  & 4.4 GHz & 2.2 GHz \\ \hline
TOPS\footnotemark[2] & 4.5E-2 & 4.5E-2  & 4.5E-2 & 4.1E-2  & 2E-2 & 9.9E-3  & 4.9E-3
 & 2.5E-3 \\ \hline
\end{tabular}
\caption{TOPS performance of hybrid optical processor with various different calculation precision levels.}
\label{Hybrid_scheme}

\begin{tablenotes}
    \item[1] The system speed is determined by the lesser of two speeds: the maximum achievable speed of the \gls*{ADC} and the input data speed through the digital I/O.
    \item[2] The convolution task utilizes a 3$\times$3 size kernel, and the weight value precision is 8 bit. The TOPS is calculated by multiplying the hardware scale (i.e., L = 9) with the speed of the system, and then dividing by the bit depth of input data, in accordance with the working principle of \gls*{HOP}.
    
\end{tablenotes}
\end{threeparttable}
\end{table}

However, it should be noted that the hybrid scheme can have some advantage over the analog scheme in terms of speed and computing power, according to equations (\ref{fs_analog}), (\ref{fs_hybrid}), (\ref{tops_analog}), and (\ref{tops_hybrid}). To quantitatively compare the two schemes, we assume a convolution task with an operator of size 3$\times$3. It is assumed that the input data 'd', transmitted through a high-speed DAC or pattern generator, has a rate of 40 GHz with precision ranging from 1 bit to 8 bit. The weight has a precision of 8 bit. Employing the equations and assuming an input data rate of 40 GHz, we can calculate the required \gls*{ENOB} of the \glspl*{ADC} (equation (\ref{enob_analog}) and (\ref{enob_hybrid})), achievable \gls*{ADC} sampling rate in theory (equation (\ref{fs_analog}) and (\ref{fs_hybrid})), and the \gls*{TOPS} (equation (\ref{tops_analog}) and (\ref{tops_hybrid})). These results are presented in Table \ref{Analog_scheme} and Table \ref{Hybrid_scheme}.

The computing power for both the analog scheme and the \gls*{HOP} scheme are further illustrated in Supplementary Figure \ref{Comparision_analog_hybrid}c. It is evident that for both the analog and hybrid schemes, the \gls*{TOPS} decreases as the calculation precision increases. Notably, an intersection occurs at 3-bit calculation precision. This indicates that the analog scheme holds an advantage in computing power at low-bit precision. However, the hybrid scheme demonstrates better performance with higher precision. 

We also explore various other input rates, including 1 GHz, 10 GHz, and 100 GHz, as shown in Supplementary Figure \ref{Comparision_analog_hybrid}a,b, and d. With higher input rates, the advantage in \gls*{TOPS} by the hybrid scheme gets bigger and the analog scheme gradually loses its advantage at low calculation precision regions. 

\begin{figure}[htbp]
    \centering
    \includegraphics[width=1\linewidth]{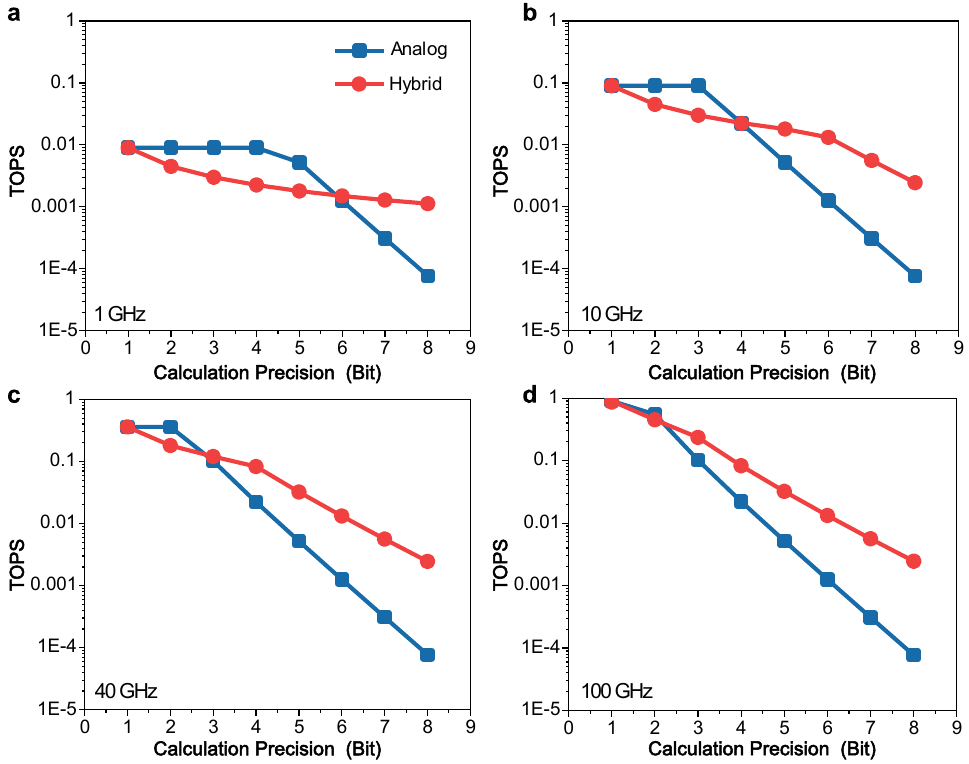}
    \caption{\textbf{The computing power of analog and hybrid schemes: a comparison across various calculation precisions (from 1-bit to 8-bit) and input data speeds (1 GHz, 10 GHz, 40 GHz and 100 GHz).}}
    \label{Comparision_analog_hybrid}
\end{figure}

\section{Detailed weight loading and \acrshort*{MRM} modulation bias tuning}
\begin{figure}[htbp]
    \centering
    \includegraphics[width=0.85\linewidth]{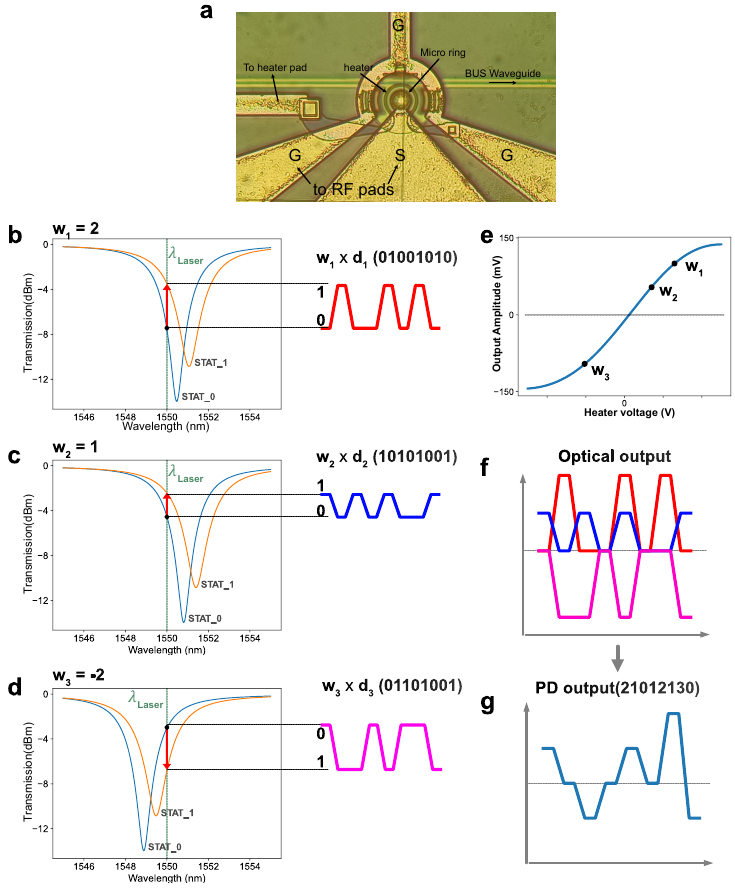}
    \caption{\textbf{Detailed weight loading and \gls*{MRM} modulation bias tuning.} \textbf{a,} Top-view microscope image of a single \gls*{MRM}. \textbf{b-d,} \gls*{MRM} working status at weight value w = 2, w = 1, and w = -2, respectively. \textbf{e,} The relation between heater voltage and amplitude of the photodetected multiplication result. \textbf{f-g,} The addition of multi-wavelength optical signals results in a multilevel digital signal at the baseband representing the inner product of the input vector $d$ and weight vector $w$. Note the zero voltage level does not necessarily correspond to the minimum level of the baseband multilevel digital signal.}
    \label{Figure_S1}
\end{figure}
The detailed weight loading scheme for the photonic multiplier based on the \gls*{MRM} is shown in Supplementary Figure \ref{Figure_S1}. We plot the working status, i.e., transmission curve, of the \gls*{MRM} when loading weight values of 2, 1, and -2, respectively, in Supplementary Figure \ref{Figure_S1}b-d. The blue and orange curves indicate the 'off' operating status (STAT$\_$0) and the 'on' operating status (STAT$\_$1) of the \gls*{MRM}. The amplitude of the output is determined by the optical intensity difference between these two statuses, which can be adjusted by tuning the applied voltage on the microheater, i.e., the working wavelength of the \gls*{MRM}. In other words, the weight of the multiplier is adjusted by tuning the voltage applied to the microheater. Positive and negative weight values can both be realized by tuning the initial working wavelength of the \gls*{MRM}. If the initial working wavelength resides on the red side (longer wavelengths) of the laser wavelength, positive weight values can be loaded. On the contrary, an initial working wavelength at the blue side of the laser wavelength will result in a flip of logic 0 and 1, thus a negative weight value is loaded. 

The relationship between the weight and the microheater bias is measured and shown in Supplementary Figure \ref{Figure_S1}d. The measurement can then be used as a lookup table for the loading of the normalized weights. The weight values in Supplementary Figure \ref{Figure_S1}a-c are illustrated accordingly. Supplementary Figure \ref{Figure_S1}e-f shows the process of photodetection, which sums up all three weighted modulations, accomplishing the calculation of the inner product of the input vector $d$ and weight vector $w$. It should be noted that the zero voltage level does not necessarily correspond to the minimum level of the baseband multilevel digital signal, as indicated in Supplementary Figure \ref{Figure_S1}f.

\section{Convolution results from alternative operators and images}
\begin{figure}[htbp]
    \centering
    \includegraphics[width=0.9\linewidth]{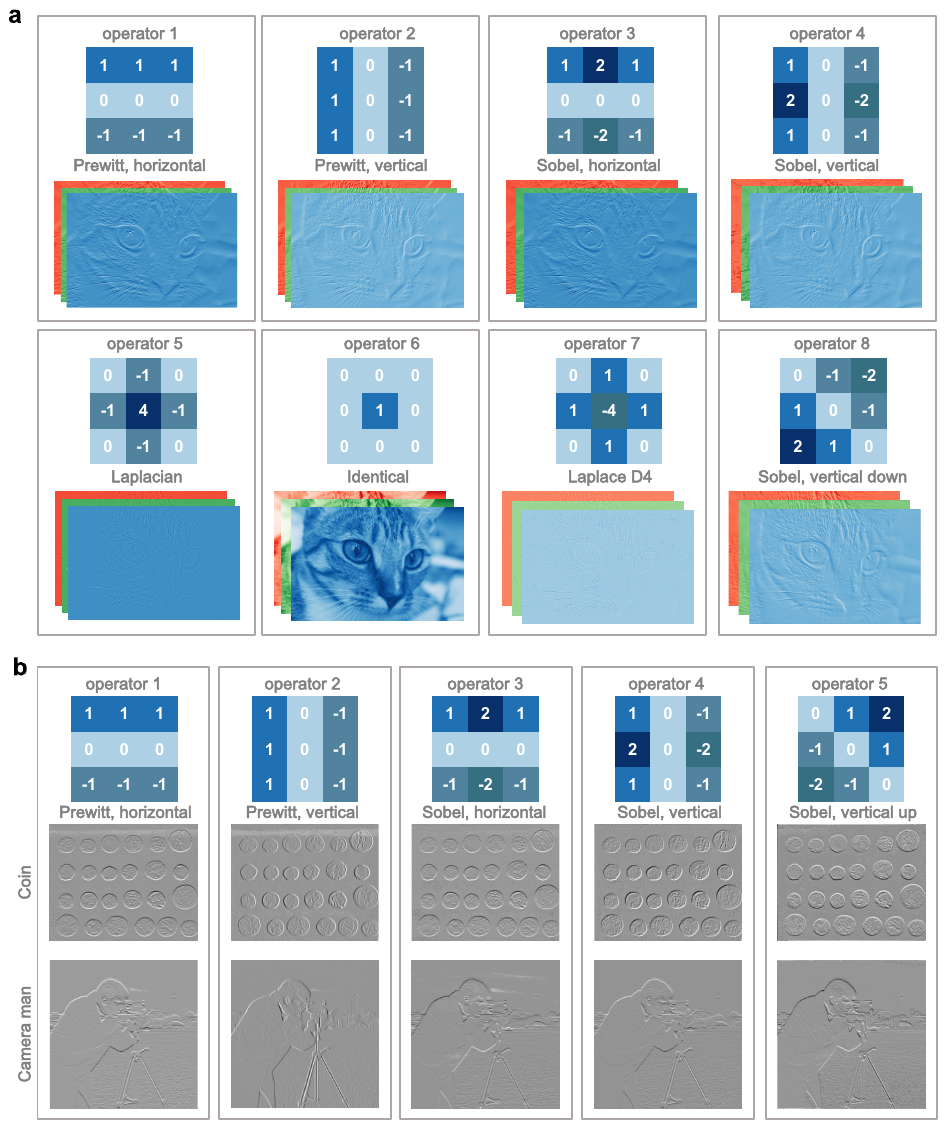}
    \caption{\textbf{Alternative convolution results.} \textbf{a,} Results using alternative convolution operators performed on an 8-bit image. \textbf{b,} Results performed on alternative images.}
    \label{Figure_S2}
\end{figure}
To further evaluate the performance of our \gls*{HOP}, we employed a variety of convolution operators and used alternative images. The convolution operators include Prewitt, Sobel, Sharpen D4, Identical, and Laplace D4. The convolutions are performed on an 8-bit image (Chelsea\cite{van2014scikit}) and the results are shown in Supplementary Figure \ref{Figure_S2}a. Results from convolutions performed on alternative images in scikit dataset\cite{van2014scikit} are shown in Supplementary Figure \ref{Figure_S2}b. The results clearly show a consistent performance with a minimized number of noisy pixels, indicating a reliable calculation precision across all tested operators and images.

\section{Convolution with variable bit precision}
\begin{figure}[htbp]
    \centering
    \includegraphics[width=0.9\linewidth]{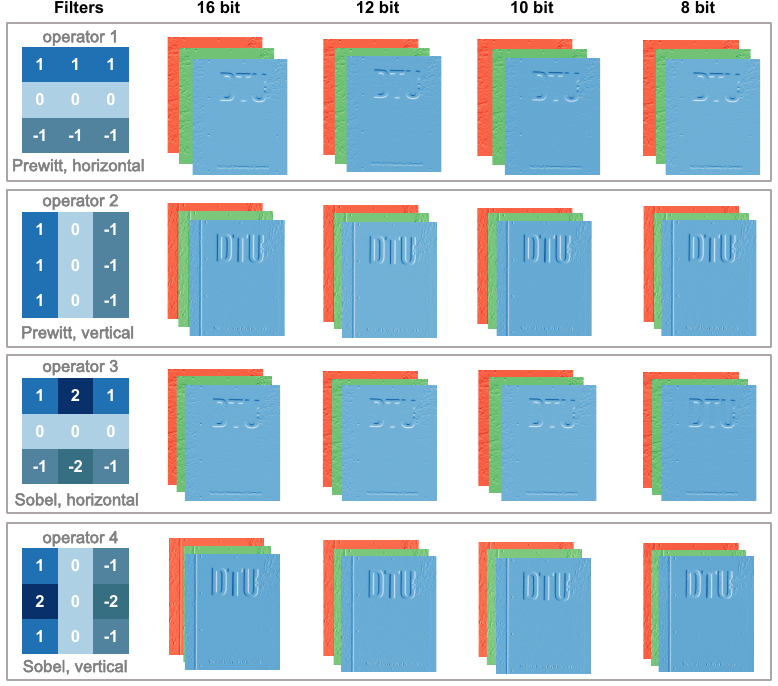}
    \caption{\textbf{Convolution performed using variable bit precision.}}
    \label{Figure_S3}
\end{figure}
High bit-depth images are essential for some neural network applications. Higher bit-depths correspond to an increased number of possible color levels, thus enabling a richer display of color information. Many standards incorporate color depths of more than 8 bits, for example, 16-bit RAW, 12-bit \gls*{DVS}, and 10-bit \gls*{HDR}. An 8-bit \gls*{SDR} image offers 2$^8$=256 distinct color gradations, while a 16-bit image provides 2$^{16}$=65536 gradations. 

To properly process such images without losing information, a calculation precision of 16 bits is required, and $M=N=16$. Even for a small scale, i.e., $L=9$ (for a 3$\times$3 convolution), the analog computing scheme requires \glspl*{DAC} with 16-bit resolution and \glspl*{ADC} with 36-bit resolution (Eq. \ref{enob_analog}). While this might be possible, it is at the expense of a much slower \glspl*{ADC} conversion rate which then becomes a bottleneck. As previously discussed, for our \gls*{HOP}, the number of signal levels of the output and the resolution requirement for the \glspl*{ADC} do not depend on the precision of the input $d$. The \gls*{HOP} scheme eliminates the use of the \glspl*{DAC} and requires \glspl*{ADC} with a much reduced resolution of 20 bits (Eq. \ref{enob_hybrid}), which is much more manageable and a significant improvement over analog computing schemes. 

Here, we demonstrate the capability of our \gls*{HOP} by processing an image with variable bit precision, using both Prewitt and Sobel operators on the same 16-bit image in the main paper, taken and preprocessed from a digital camera. The results are shown in Supplementary Figure \ref{Figure_S3}. No performance difference is observed across images with different bit depths. This demonstrates that our \gls*{HOP} could indeed work with input values of any precision without sacrificing performance.

\section{Scalability and system comparisons}

Deep learning tasks usually require a greater number of convolution kernels and large matrix multiplication scales that are beyond the capability of current \glspl*{ONN}. A large number of wavelength channels for the multi-wavelength light source together with massive hardware scaling for the \gls*{PIC} is required. The multi-wavelength light source can be an optical frequency comb that provides more than a hundred wavelength channels and can be co-integrated to the \gls*{PIC}\cite{kong22np}.

The \gls*{MRM} based \gls*{HOP} has the advantage in terms of dimensions compared to \gls*{MZI} based \glspl*{ONN}\cite{shen17NP}. Envisioning a matrix multiplication task for $64\times64$ matrices, 4096 \glspl*{MRM} are required for the \gls*{HOP} scheme. Taking into account the typical feature size (\SI{15}{\micro\meter} in diameter) and the pitch (\SI{30}{\micro\meter}), the required chip size is estimated to be \SI{3.69}{\milli\meter}$^2$, a size that is feasible for manufacturing. Achieving a high yield for a chip on this scale is very challenging. However, the \gls*{MRM} based \gls*{HOP} can introduce redundancies by implementing an excess number of \glspl*{MRM} for each convolution operator and an excess number of \gls*{MRM} arrays. Thanks to the nature of the \glspl*{MRM}, one down-performed \gls*{MRM} can be replaced by another \gls*{MRM}. And one \gls*{MRM} array can also be substituted by another array. 

Simultaneously operating a large number of \glspl*{MRM} also presents a significant challenge, particularly due to discrepancies between design and manufacturing. It often requires correction by the microheater modulator to align the working wavelength of the \glspl*{MRM} to the multiwavelength light source, which could consume a significant amount of power. However, the post-fabrication trimming technique could solve this power consumption\cite{yu22trimming}.

Given the above discussion, the high noise tolerance of our \gls*{HOP}, the removal of \glspl*{DAC}, and the relaxed resolution requirement for the \glspl*{ADC}, the \gls*{MRM} based \gls*{HOP} can be manufactured for large-scale neural network applications.

Using achievable parameters for the photonic devices, we have compared our \gls*{HOP} with a state-of-the-art analog \gls*{ONN} processor \cite{feldmann21Nature} and Google's \gls*{TPU}\cite{jouppi17} in terms of compute density ($\text{TOPS}/\text{mm}^2$) and calculation precision. The results are shown in Supplementary Table \ref{tab:comparison}. 

The \gls*{HOP} scheme in this study operates at a clock speed of 7.5 GHz, processing matrices of size 3x3 and with a computing capacity of 8.44 GOPS. The effective area occupied by each \gls*{MRM} amounts to \SI{900}{\micro\meter}$^2$. The diameter of the ring is \SI{15}{\micro\meter}, and the pitch of the \glspl*{MRM} is \SI{250}{\micro\meter} in our fabricated chip, considering design redundancy\cite{kong2020intra}. Yet, the pitch could be reduced to \SI{30}{\micro\meter} or less if techniques such as undercut is used to manage thermal crosstalk\cite{dong2010low}. This yields a compute density of \SI{1.04}{TOPS\per\milli\meter}$^2$. The computing power can potentially be scaled up by increasing the operating speed to more than \SI{40}{\giga\hertz} due to the elimination of \gls*{DAC} and released resolution requirement for \glspl*{ADC}. As a result, the \gls*{HOP} can potentially achieve a compute density larger than \SI{5.6}{TOPS\per\milli\meter}$^2$.

Evaluating the power consumption of the computing system requires knowledge of the power consumption from all supporting electronics, particularly the signal format converters (i.e., \glspl*{DAC} and \glspl*{ADC}). However, this type of power consumption is often overlooked in the literature discussing analog \glspl*{ONN}. And it could depend on many factors such as the fabrication technology node. Therefore, we choose to eliminate the comparison of power consumption here. However, it should be noted the power consumption of our \gls*{HOP} can be much reduced compared with analog \glspl*{ONN} due to the removal of \glspl*{DAC} and the released resolution requirement for the \glspl*{ADC}. Nevertheless, the power consumption analysis for \gls*{ONN} lacks standards and should be investigated extensively in future research.

\begin{table}[ht]
\centering
\begin{tabular}{|C{4cm}|C{1cm}|C{3.3 cm}|C{1.6cm}|}
\hline
\rowcolor{Gainsboro!300}
Technology & Clock speed &  Compute Density (TOPS/mm$^2$) & Precision (bits) \\ \hline


\rowcolor{Gainsboro!100}
Analog optical neural network\cite{feldmann21Nature}  & 12 GHz  & 1.2  & 5 \\ \hline

Google TPU (ASIC)\cite{jouppi17} & 700 MHz & 0.28 & 8 \\ \hline

\rowcolor{Gainsboro!100}
GPU(Nvidia H100 PCIe) & 1.62 GHz & 3.7 & 8 \\ \hline

\gls*{HOP} & 7.5 GHz & 1.04 (3x3 matrix) & 16 \\ \hline

\rowcolor{Gainsboro!100}
\gls*{HOP} & 40 GHz & 5.6 (64x64 matrix) & 16 \\ \hline



\end{tabular}
\caption{Performance comparison between different neural network processor schemes. Note that the computations here are fixed-point operations.}
\label{tab:comparison}

\end{table}
